\documentclass[sn-mathphys-num]{sn-jnl}

\usepackage{graphicx}%
\usepackage{multirow}%
\usepackage{amsmath,amssymb,amsfonts}%
\usepackage{amsthm}%
\usepackage{mathrsfs}%
\usepackage[title]{appendix}%
\usepackage{xcolor}%
\usepackage{textcomp}%
\usepackage{manyfoot}%
\usepackage{booktabs}%
\usepackage{algorithm}%
\usepackage{algorithmicx}%
\usepackage{algpseudocode}%
\usepackage{listings}%
\usepackage{kotex} 

\theoremstyle{thmstyleone}%
\theoremstyle{thmstyletwo}%

\theoremstyle{thmstylethree}%

\begin{document}

\title[Article Title]{Power of Cooperative Supervision: Multiple Teachers Framework for Enhanced 3D Semi-Supervised Object Detection}


\author*[1]{\fnm{Jin-Hee} \sur{Lee}}\email{jhlee07@dgist.ac.kr}
\equalcont{These authors contributed equally to this work.}

\author[2]{\fnm{Jae-Keun} \sur{Lee}}\email{lejk8104@gmail.com}
\equalcont{These authors contributed equally to this work.}

\author[1]{\fnm{Je-Seok} \sur{Kim}}\email{jeseok@dgist.ac.kr}

\author*[1,2]{\fnm{Soon} \sur{Kwon}}\email{soonyk@dgist.ac.kr}

\affil[1]{\orgdiv{Division of Automotive}, \orgname{DGIST}, \orgaddress{\city{Daegu}, \country{Republic of Korea}}}

\affil[2]{\orgname{FutureDrive Inc.}, \orgaddress{\city{Daegu}, \country{Republic of Korea}}}


\abstract{To ensure safe urban driving for autonomous platforms, it is crucial not only to develop high-performance object detection techniques but also to establish a diverse and representative dataset that captures various urban environments and object characteristics. To address these two issues, we have constructed a multi-class 3D LiDAR dataset reflecting diverse urban environments and object characteristics, and developed a robust 3D semi-supervised object detection (SSOD) based on a multiple teachers framework. This SSOD framework categorizes similar classes and assigns specialized teachers to each category. Through collaborative supervision among these category-specialized teachers, the student network becomes increasingly proficient, leading to a highly effective object detector. We propose a simple yet effective augmentation technique, Pie-based Point Compensating Augmentation (PieAug), to enable the teacher network to generate high-quality pseudo-labels. Extensive experiments on the WOD, KITTI, and our datasets validate the effectiveness of our proposed method and the quality of our dataset. Experimental results demonstrate that our approach consistently outperforms existing state-of-the-art 3D semi-supervised object detection methods across all datasets. We plan to release our multi-class LiDAR dataset and the source code available on our Github\footnote {https://github.com/JH-Research/MultipleTeachers} repository in the near future.}

\keywords{Semi-supervised learning, Object detection, Autonomous driving}



\maketitle

\section{Introduction}\label{sec1}
For autonomous vehicles to drive safely in urban environments, it is crucial to accurately recognize and understand the complex urban surroundings. Therefore, LiDAR-based 3D object detection, which provides robust distance measurements essential for these tasks, has attracted significant attention from research groups. However, developing high-performance detectors that leverage deep learning requires training on diverse road environments. Thus, many researchers often rely on substantial amounts of labeled data. Despite efforts to acquire extensive labeled datasets, the time and cost considerations make building large-scale 3D datasets for autonomous driving particularly challenging. 

Unlike image-based datasets \cite{bib1, bib2}, LiDAR-based datasets \cite{bib3, bib4, bib5} are not only fewer in number but also more demanding in terms of collecting and labeling data. Gathering data using vehicle systems equipped with LiDAR sensors under various environmental conditions (e.g., weather, road complexity) requires significantly higher costs and efforts. This is because 3D LiDAR point clouds is more challenging due to the lack color differentiation and the sparse nature of the data, which even is hard to distinguish object sizes and orientations.

Due to the irregular and sparse characteristics of LiDAR point data, object detection is more challenging compared to camera image data. However, LiDAR provides highly accurate 3D distance information and can detect objects in a 360-degree field of view with a single sensor, eliminating the need for additional data analysis techniques such as image-based feature matching. As a result, object detection using LiDAR has been a consistent focus of research, with numerous state-of-the-art models \cite{bib6, bib7, bib8, bib9, bib10} being developed.

To effectively train on LiDAR point clouds, 3D object detectors generally divided into two categories: voxel-based and pillar-based methods. Both approaches begin by converting point clouds into uniform voxel or pillar representations. These transformed data are then processed through sequential backbone, neck, and head modules to predict 3D objects. These approaches typically convert point clouds into grid structures, which make it possible to apply convolutional neural networks (CNNs). However, this grid-based method can lead to the inevitable loss of important point information. This negatively impacts detection accuracy \cite{bib23}. Particularly for real-time processing in autonomous vehicles, pillar-based methods, which use larger grid sizes, tend to have lower accuracy compared to voxel-based methods.

To enhance the performance of the detector, this paper employs a 2-stage voxel-based approach to address the aforementioned drawbacks. In this approach, 3D bounding boxes predicted in the first stage are used to pool region of interest (ROI) features from the voxel or point features, generating new ROI features. By leveraging these features during object detection, our model can predict more accurately positioned bounding boxes. However, relying solely on labeled data for object detection has its limitations. Hence, there has been growing attention towards 3D SSOD techniques \cite{bib11, bib12, bib13, bib14, bib15}, which aim to improve detection performance by training on a small amount of labeled data alongside a large volume of unlabeled data. Inspired by the pioneering MeanTeacher model \cite{bib16}, these studies adopt a teacher-student framework. In this framework, the teacher network generates pseudo-labels, and the student network learns from both these pseudo-labels and the labeled data, gradually enhancing object detection accuracy. Although these approaches being proven effective, several important challenges that need to be addressed.

Firstly, the quality of pseudo-labels generated by the teacher network significantly influences the accuracy of object detection. Therefore, generating reliable pseudo-labels is paramount. As the quality of generated pseudo-labels greatly affects the accuracy of the student network, most existing methods focus solely on developing effective pseudo-label generators by utilizing the detection results of a single teacher model. However, since objects vary significantly in aspect ratio and size across different categories, such as vehicles and pedestrians, relying on a single model to generate pseudo-labels without considering the distinct characteristics of each category has clear limitations in improving detection accuracy. One approach to address this issue is to group objects based on similarities in characteristics such as size across classes and design a more specialized teacher network accordingly.

Secondly, there is a lack of strong augmentation techniques to enhance the generalization ability of the student network. Current 3D SSOD research primarily relies on weak augmentation techniques such as flip, scaling, and rotation. While these techniques may slightly alter the distribution of point clouds, they fail to dramatically augment the diversity of ground truths (GT). Therefore, the student network may be restricted in effectively learning from a diverse of scenes. This is particularly challenging when dealing with scenarios where do not have sufficient points or occluded by other objects, making accurate predictions significantly difficult. Existing studies have not only adequately addressed these issues but also, highlighting the increasing need for research on strong augmentation techniques that can overcome these limitations.

\begin{figure}[t]
\centering
\includegraphics[width=1.0\textwidth]{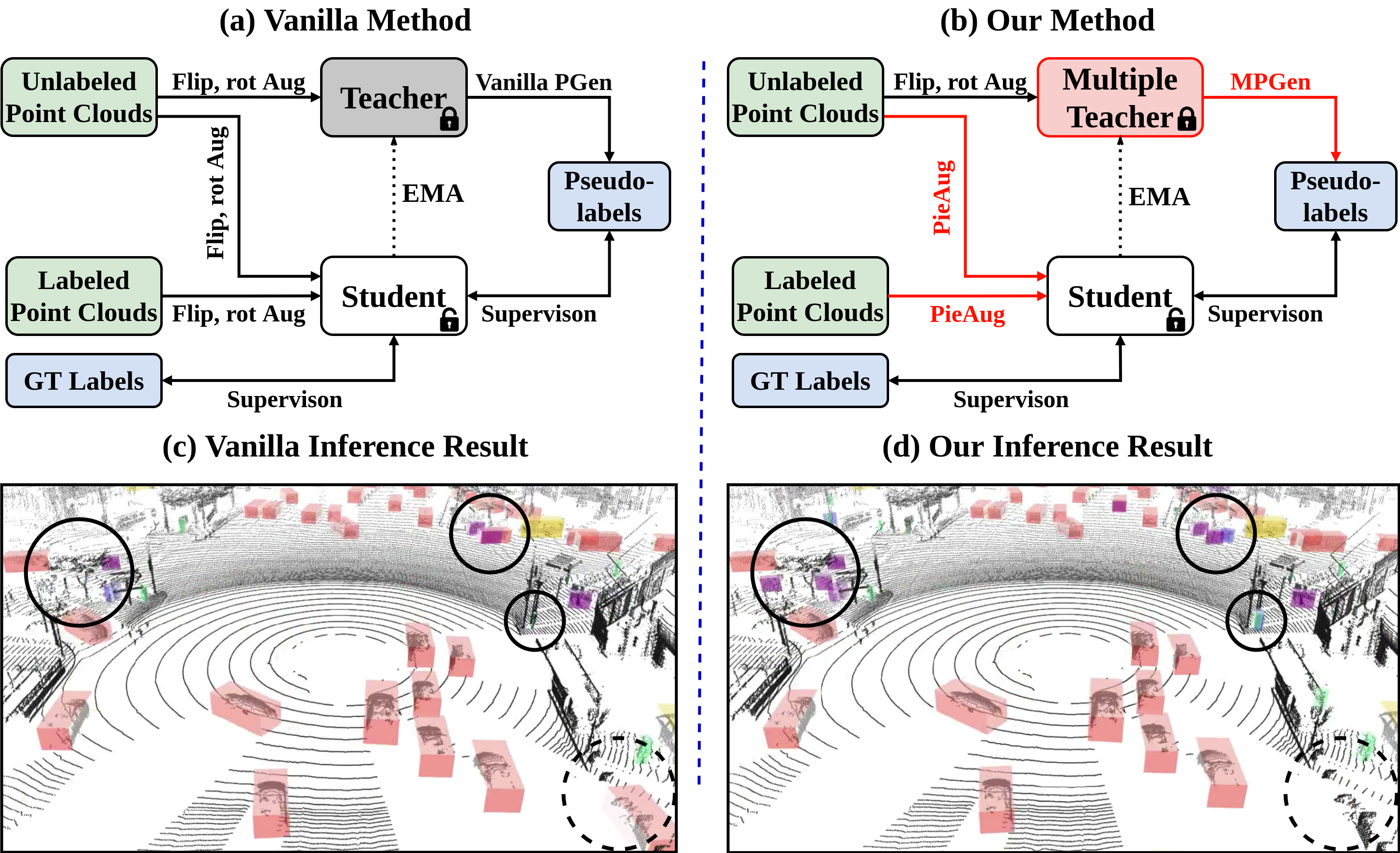}
\caption{\textbf{Comparisons between the previous method and our MultipleTeachers framework.} (a) utilizes a single teacher-student paradigm to generate pseudo-labels with flip and rotation augmentations. In contrast, (b) employs the multiple teachers-student paradigm to generate more precise pseudo-labels using the novel strong augmentation, PieAug. (c) represents the 3D detection results predicted by the vanilla 3D SSOD method. (d) indicates the 3D detection results predicted by our proposed MultipleTeachers. Note that, `PGen' refers to pseudo-label generator, and `MPGen' denotes multiple teachers-based pseudo-labels generator}\label{fig1}
\end{figure}

In this paper, we propose a novel 3DSSOD framework called MultipleTeachers, designed to address the aforementioned issues by creating specialized teachers network for various categories and having these multiple teachers collaboratively guide the student network effectively. To this end, we focus on three groups: vehicles, pedestrians, and cyclist, each represented by the specialized teacher. This framework leverages high-quality pseudo-labels generated by category-specific expert teachers to maximize the learning efficiency of the student network, thereby significantly enhancing object detection capabilities.

One of the key advantages of MultipleTeachers is that each teacher specializes in different categories of objects, enabling the generation of more accurate pseudo-labels for objects with varying aspect ratios and sizes. This effectively addresses the issue of low-quality pseudo-labels commonly encountered in traditional single teacher network paradigms. In addition, to mitigate issues caused by the sparsity characteristics of point clouds, we propose a novel data augmentation technique called Pie-based point compensating augmentation (PieAug). This technique divides the point cloud into several uniformly sized pies and analyzes the distribution of objects within each pie. It is designed to compensate for the points in sparse pie regions with points from other dense pie regions. This approach provides richer training information for objects located far from the LiDAR sensor or obscured by other objects, thereby enhancing the generalization ability of the student network and significantly improving detection performance.

Furthermore, to accelerate advancements in autonomous driving research, we recognize the necessity of not only designing high-performance detectors but also establishing the dataset for autonomous driving. Consequently, we introduce the LiDAR Open Dataset (LiO), which comprehensively reflects diverse urban environments, intersections, traffic situations, and dedicated roadways. This dataset includes various classes of objects such as vehicles, buses, trucks, other vehicles, pedestrians, motorcycles, and bicycles. The LiO dataset ensures data balance across these classes and undergoes strict refinement and meticulous annotation to achieve high-quality data.

To demonstrate the superiority and effectiveness of MultipleTeachers, extensive experiments are conducted on three large-scale autonomous driving datasets: Waymo Open Dataset (WOD) \cite{bib3}, KITTI \cite{bib4}, and LiO. The experimental results show that our proposed detector achieved state-of-the-art performance surpassing existing 3D SSOD models on all three datasets. Particularly, with only 15\% labeled and 85\% unlabeled data, our detector achieves 71.7 mAP on the KITTI dataset and 57.8 mAP on the LiO dataset. Moreover, on the WOD dataset, using just 1\% labeled and 99\% unlabeled data, our method achieves a significant performance improvements compared to previous state-of-the-art detectors, reaching 41.1 mAP (Level 2).

Finally, when comparing the object detection results of the traditional vanilla 3D SSOD method with our detector, the MultipleTeachers approach shows markedly superior performance. As highlighted by the solid circles in Fig. \ref{fig1} (c) and (d), our method more accurately predicts the bounding box positions of pedestrians and cyclists. In addition, as observed in the dashed circles, our detector exhibits relatively fewer misclassifications of vehicles compared to the vanilla method. This improvement can be attributed to the high-quality pseudo-labels generated through the cooperative process among multiple teachers proposed in this paper, which plays a crucial role in driving the strong detection performance of the student network.

\section{Related Work}\label{sec2}
\subsection{3D Supervised Object Detection}\label{subsec2.1}
The technology of 3D object detection has evolved in various methods to effectively process point clouds, which have irregular and sparse characteristics. Specifically, it can be broadly categorized into point-based and grid-based approaches, depending on how they handle the points.

First, the point-based approaches \cite{bib17, bib18, bib26} directly extract features from raw point clouds. Notable examples include PointNet \cite{bib17} and PointNet++ \cite{bib18}. Since these methods connect raw points directly to a CNN for feature extraction, they have the advantage of minimizing point information loss and preserving the spatial structure information among points. However, they also introduce additional overhead due to the resource costs associated with point sampling and grouping processes, which can impact real-time processing.

The grid-based approach can be further subdivided into voxel-based \cite{bib20, bib21, bib22, bib6, bib8, bib10, bib37, bib38} methods and pillar-based \cite{bib19, bib7, bib9, bib35, bib36}. One of the early models in the voxel-based approach, VoxelNet \cite{bib20}, uses a simple PointNet \cite{bib17} to transform irregular point clouds into regular voxel volumes. Then, it applies voxelized point clouds to a conventional 3D CNN backbone network to extract features, proposing an end-to-end learning method. SECOND \cite{bib21} introduces sparse convolutional neural networks (SparseCNNs) \cite{bib25} to alleviate the geometrically increasing computational complexity issue associated with the increase in input resolution in VoxelNet. By avoiding unnecessary computations in empty voxels, SparseCNNs enhance the efficiency of the backbone network. On the other hand, pillar-based object detectors like PointPillar \cite{bib19}, which prioritize real-time processing capabilities, encode point clouds into simple pillar shapes with a fixed height, without considering the height dimension of the point clouds. These pillar-encoded data are then processed by a 2D CNN to encode bird eye view (BEV) features necessary for 3D box prediction. While recent grid-based methods generally achieve superior performance, the process of gridding points inevitably leads to the loss of critical point information, which can negatively impact the performance of the object detector. 

In recent object detection research, the performance of detectors has been improved by introducing center-based methods \cite{bib30, bib6, bib7, bib8, bib9, bib10, bib35, bib36, bib37, bib38}, which directly predict the centers of objects instead of relying on the use of pre-defined anchor boxes commonly used in previous approaches. These methods are designed to more accurately predict the position and size of 3D bounding boxes based on the center points of objects, typically employing a 1-stage detection approach. Additionally, these models have proposed various architectures to enhance detection performance. AFDetV2 \cite{bib6} improves detection performance by applying self-calibrated convolution to the neck module composed of a region proposal network (RPN) to more effectively encode BEV features. PillarNet \cite{bib7} and PillarNeXt \cite{bib9} introduce backbone modules based on 2DSparseCNNs to more effectively encode pillar information. Re-VoxelDet \cite{bib10} improves key components of CenterPoint \cite{bib30}, such as backbone, neck, and header, and proposes an effective neck structure that can hierarchically integrate 3D raw voxel features.

Another approach to more accurate 3D object detection is the extension of 2-stage RCNN-based 2D object detection methods \cite{bib34} to 3D \cite{bib26,bib22,bib23,bib24}. PointRCNN \cite{bib26} directly extracts 3D ROIs from point clouds using PointNet++ \cite{bib18} as the encoding layer. Subsequently, in the 2-stage network, ROI pooling is performed to fuse semantic feature information of pooled points with predicted boxes to achieve superior object detection performance. Voxel-RCNN \cite{bib22} enables more accurate box prediction by applying the Voxel-ROI Grid Pooling module, which aggregates coarse voxel features to generate ROIs. PV-RCNN \cite{bib23} preserves object position and spatial information by integrating various scales of 3D sparse features extracted from a 3D sparse backbone and raw point information, combining the advantages of voxel and point methods for more accurate object detection.

For a fair comparison with existing methods, our proposed MultipleTeachers framework adopts PV-RCNN \cite{bib23} as the baseline detector. However, our framework is compatible with various 1-stage and 2-stage detectors. Detailed comparisons experiments on the performance of various detectors are presented in Table~\ref{tab10} in Sec.~\ref{sec4.3.5} of this paper.

\subsection{3D Semi-Supervised Object Detection}\label{subsec2.2}
Recently, in the 2D image domain, research on semi-supervised learning (SSL) \cite{bib16, bib39, bib40} has been actively pursued to improve the performance of deep learning-based models by leveraging not only a small amount of labeled images but also a large number of unlabeled images. Motivated by the significant success of SSL research, many researchers have attempted to apply this technique to the field of 3D object detection. Most 3D SSOD methods adopt the widely used teacher-student paradigm \cite{bib16, bib27} from 2D SSOD. In this framework, the student network is trained with a large scale of pseudo-labels generated from the teacher network. The weights of the teacher network are updated using an exponential moving average (EMA) of latest weights of the student network, facilitating interaction between the teacher and student networks, which can significantly impact performance.

In general, generating accurate 3D bounding boxes from LiDAR data requires considering 9 degrees-of-freedom (DoF), making it more challenging compared to 2D bounding boxes in images. This is a particularly crucial issue in the field of 3D semi-supervised object detection, where generating a large quantity of high-quality pseudo-labels is essential. Therefore, various approaches have been proposed in recent research to develop effective pseudo-labels generators for this purpose \cite{bib11, bib12, bib13, bib14, bib15}.

3DIoUMatch \cite{bib11} proposes a multiple fixed threshold strategy that considers both the classification score and the accuracy of box positions by utilizing the intersection of union (IoU) estimation score to effectively filter out pseudo-labels with low-quality while retaining high-quality ones. DDS3D \cite{bib13} suggests a strategy based on dynamic thresholds to increase the diversity of pseudo-labels provided to the student network. This approach assumes the incompleteness of pseudo-label quality generated by the teacher network initially, gradually improving the quality of pseudo-labels as the training progresses to ensure the stability of student network learning and enhance performance. DetMatch \cite{bib12} proposes a multi-modal-based pseudo-labels generation process that integrates detection results from images and 3D detection results from point clouds to improve performance. HSSDA \cite{bib15} defines dual thresholds by matching predicted and ground truth boxes from labeled data and applies these thresholds to hierarchically generate pseudo-labels. This strategy optimizes student network learning and enhances detection performance by providing guidance of various difficulty levels to the student network based on the confidence of pseudo-labels.

As mentioned above, existing research has focused on designing threshold-based filtering techniques to suppress False Negative (FN) and False Positive (FP) included in pseudo-labels, aiming to generate more accurate pseudo-labels.

\subsection{Data Augmentation based on 3D Point Clouds}\label{subsec2.3}
Research on enhancing the geometric diversity of point clouds collected from LiDAR is demonstrated in various studies \cite{bib21, bib28}. PolarMix \cite{bib28} is an augmentation technique applied in the field of semantic segmentation based on the polar coordinates of LiDAR. This technique increases data diversity by extracting point clouds from two different scenes and mixing techniques such as swapping, global rotation, and copy-paste between point clouds. GT copy-augmentation in SECOND \cite{bib21} is one of the strong augmentation techniques for 3D object detection. It significantly increases data diversity by replicating GT boxes and foreground points in new frames using a pre-defined GT box database. 

Increasing the diversity of point clouds and pseudo-labels during training is crucial for enhancing the robustness and accuracy of deep learning models. Accordingly, research on data augmentation techniques in the field of 3D SSOD \cite{bib29, bib15} is actively progressing. PseudoAugment \cite{bib29} represents the first application of the GT copy-augmentation technique to augment pseudo-labels. This method introduces three additional pseudo-labels augmentation strategies to enhance the diversity of pseudo-labels. On the other hand, HSSDA \cite{bib15} introduces a new strong augmentation technique called shuffle data augmentation to 3D SSOD. This method maximizes geometric transformations of data by dividing point clouds into N$\times$N patches and randomly shuffling these patches, enabling the generation of more diverse pseudo-labels.

In our MultipleTeachers framework, we introduce a new augmentation technique called PieAug to supplement the sparse point information of objects located in sparse regions, considering the sparsity of LiDAR point clouds. Through this approach, our detector learns from augmented point cloud data that provides rich information for objects obscured by other objects or located relatively far away. Ultimately, this enables our detector to achieve robust detection for objects occluded by other objects or situated at relatively long distances.

\section{Our Method}\label{sec3}

\subsection{Problem Definition}\label{sec3.1}
The primary objective of 3D semi-supervised object detection is to achieve superior 3D object detection performance by leveraging both labeled and unlabeled datasets during training, compared to using only labeled data. Here, we define the labeled dataset as $\mathit{\hat{D}}$=$\{\hat{p}_{n}, \hat{y}_{n}\}_{n=1}^{N^L}$ and the unlabeled dataset as $\mathit{\bar{D}}$=$\{\bar{p}_{n}\}_{n=1}^{N^U}$. Furthermore, $\mathit{p}_{n} \in \mathbb{R}^{\{3+\gamma\}}$ represents the point clouds of the $n$-th frame, which includes the 3D coordinates of the points $(x, y, z)$ and additional information $\gamma$ (e.g., intensity, ring, etc.). For the labeled data, $\mathit{\hat{y}_{n}=\{b_{i}, c_{i}\}_{i=1}^{M}}$ represents the $M$ annotation information corresponding to each point cloud, which includes bounding boxes $b_{i}$ and their respective classes $c_{i}$. During the training process, the student network learns from the large-scale pseudo-labels generated by the teacher network. This allows the student network to effectively learn feature information inherent in the unlabeled data, thereby progressively improving detection performance.

\begin{figure}[t]
\centering
\includegraphics[width=1.0\textwidth]{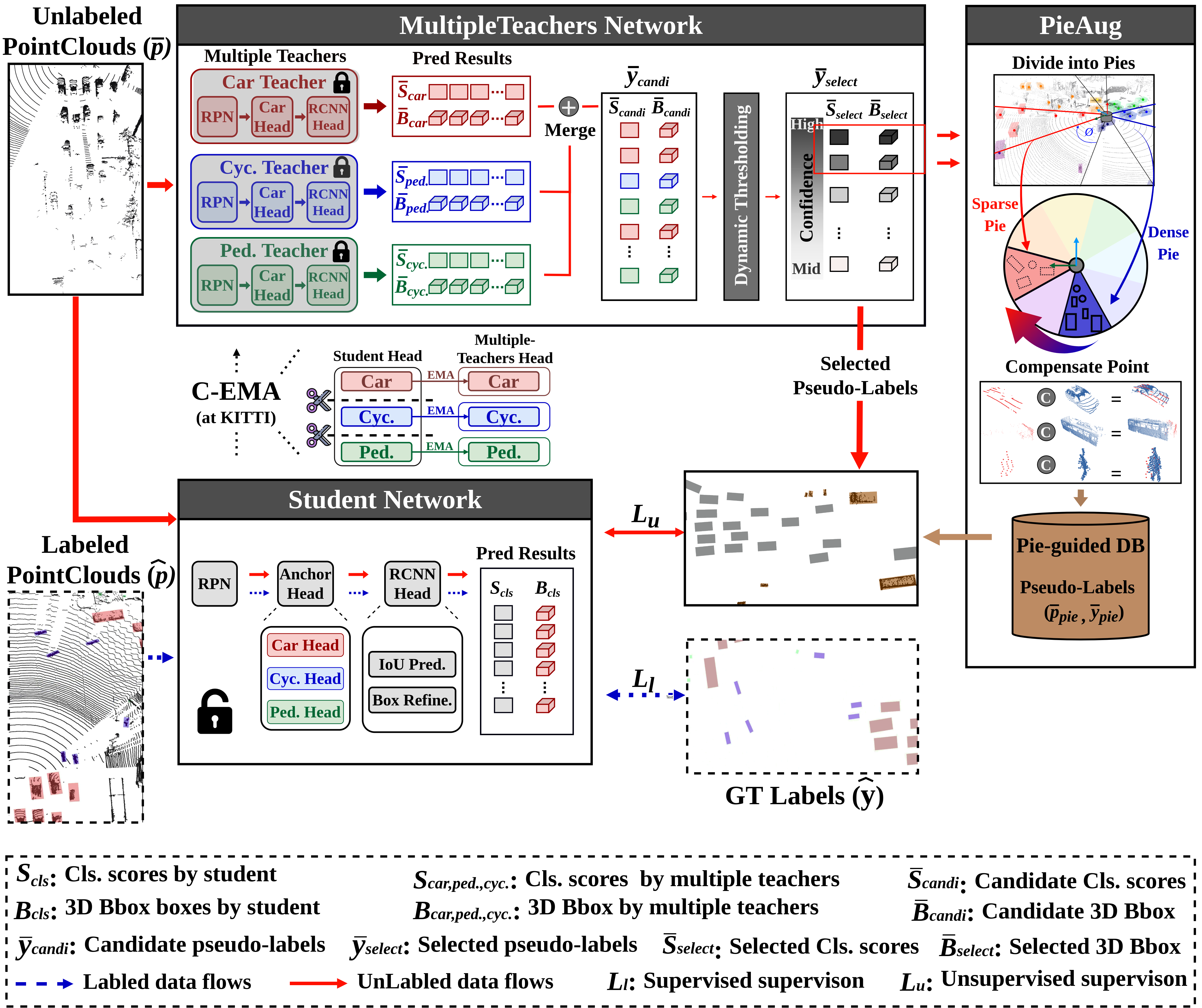}
\caption{\textbf{The overview of MultipleTeachers Framework for 3D semi-supervised object detection.} This framework consists of multiple teachers network and a single student network. Each teacher network classifies objects with different aspect ratios and shapes into specific categories, enabling more accurate predictions. By leveraging the collaboration of these specialized teacher networks per category, high-accuracy pseudo-labels are generated to train the student network. The student network not only utilizes these high-quality pseudo-labels for training but also applies a novel data augmentation strategy, named PieAug, which effectively detects occluded or distant objects from LiDAR.}\label{fig2}
\end{figure}

\subsection{Overview}\label{sec3.2}
The overall pipeline of the proposed MultipleTeachers framework is illustrated in Fig.~\ref{fig2}. This framework is designed with an innovative multiple teachers-student network structure, replacing the traditional single teacher-student network structure commonly used in 3D SSOD approaches \cite{bib11, bib12, bib13, bib14, bib15}. Our approach leverages the pseudo-labels generated by multiple teachers to train the student network, enabling it to achieve superior detection performance. Both the multiple teachers and the student network use high-performance 2-stage 3D object detectors such as PV-RCNN \cite{bib23} as the baseline model. Our framework progresses through three stages: Burn-In, Multiple Teachers Fine-tuning, and Mutual Learning, allowing the student network to gradually train meaningful feature information from large-scale unlabeled data. Each training stage is detailed in Sec.~\ref{sec3.3}. Additionally, to further enhance the detection performance of the student network during the Mutual Learning stage, we introduce a novel Pie-based point compensating augmentation, which is comprehensively described in Sec.~\ref{sec3.4}.

\begin{figure}[t]
\centering
\includegraphics[width=0.83\textwidth]{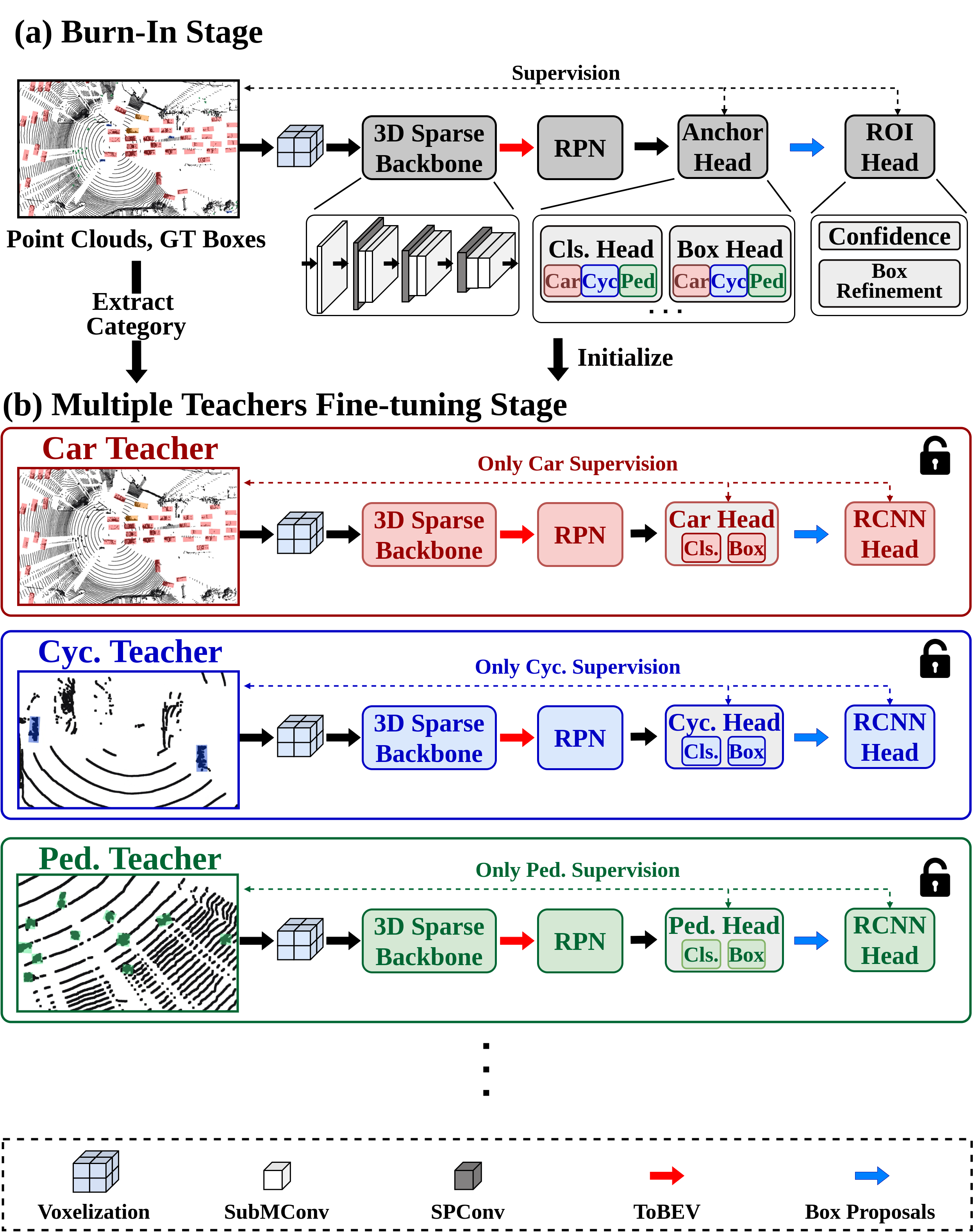}
\caption{\textbf{2-stage training process for MultipleTeachers framework.} In stage (a), the baseline detector is pre-trained with labeled data. Consecutively, the parameters and structure of this pre-trained detector are duplicated across three specialized teacher networks: Car Teacher for car categories, Ped. Teacher for pedestrian categories, and Cyc. Teacher for cyclist categories, respectively. In stage (b), each teacher network fine-tunes its category-specific parameters and the corresponding heads (i.e., Car Head, Ped. Head, Cyc. Head) to more accurately predict pseudo-labels within the designated category. This process ensures the generation of more precise pseudo-labels by each teacher}\label{fig3}
\end{figure}

\subsection{MultipleTeachers Framework}\label{sec3.3}
In general, in 3D SSOD, the object detection accuracy of the student network is heavily influenced by the quality of the pseudo-labels ($\mathit{\bar{y}}$) generated by the teacher network. Therefore, it is crucial for the teacher network to generate increasingly accurate pseudo-labels to effectively guide the student network. Traditional 3D SSOD methods \cite{bib11, bib13, bib14} typically employ a threshold strategy, where the teacher network selects pseudo-labels with high confidence and filters out low-confidence candidates. This approach helps in filtering out inaccurate candidates from the predictions made by the teacher network, thereby contributing to the generation of accurate pseudo-labels for training the student network. However, in the field of autonomous driving, which requires detecting a variety of objects in complex urban environments, it is essential to improve the detection capability for each object by learning the distinctive features such as aspect ratios and sizes. The previously developed methods struggle to address this challenge effectively.

This study aims to develop a high-performance 3D detector capable of accurately categorizing and identifying objects with various aspect ratios and shapes in autonomous driving environments. To achieve this, we employ multiple expert teachers, each specializing in specific categories, to work collaboratively and enhance detection capabilities. This approach culminates in the development of the MultipleTeachers framework, which harnesses the collective expertise of these specialized teachers to maximize detection performance.

We propose the Multiple Teachers-Guided Pseudo-Label Generator (MPGen), which leverages the expertise of category-specific teachers to generate high-quality pseudo-labels. By organizing teacher networks proficient in each category, we ensure that the pseudo-labels accurately reflect the unique characteristics of each category. The first category includes the Car Teacher, specializing in predicting objects such as cars, buses, trucks, and special vehicles, which have relatively larger aspect ratios and are sensitive to heading changes. The second category is managed by the Pedestrian Teacher (Ped. Teacher), focusing on the detection of small objects such as pedestrians. The third category involves the Cyclist Teacher (Cyc. Teacher), responsible for detecting cyclist such as motorcyclist and bicyclist.

The design of these specialized teachers for each category promotes close collaboration among the expert teachers, facilitating the generation of accurate pseudo-labels that reflect the detailed information of each category. The pseudo-labels generated through this method provide high positional accuracy and reliability for both large objects, such as cars, buses, and trucks, as well as small objects like pedestrians, bicycles, and motorcycles (see Fig.~\ref{fig6} and Fig.~\ref{fig7}).

These results demonstrate that our method overcomes the limitations of single-teacher models, which fail to sufficiently capture category-specific distinctions. By training our model with these superior pseudo-labels, we significantly enhance the accuracy and reliability of 3D object detection.

\noindent \textbf{Burn-In Stage}.
The Burn-In Stage precedes the Fine-tuning Stage of MultipleTeachers, serving as a warm-up phase where the detector is trained using only a small portion of available labeled data. As depicted in Fig.~\ref{fig3} (a), we first pre-train the baseline detector, PV-RCNN~\cite{bib23}, using the labeled data $\hat{D}$. This detector takes the point cloud $\mathit{\hat{p}}$ and its corresponding label data $\mathit{\hat{y}}$ as inputs, followed by a voxelization process to convert $\mathit{\hat{p}}$ into voxel volume. The voxelized data feeds into a 3D Sparse Backbone process, where it is progressively encoded at down-sampling ratios of \{1/1, 1/2, 1/4, 1/8\}. The multi-scale voxel features $V_{f}=$ \{$V_1$, $V_2$, $V_3$, $V_4$\} generated in this process have channel sizes of \{16, 32, 64, 64\}, respectively. 

The features extracted from the last stage of the backbone, $V_4$, are projected onto the BEV plane and passed through the RPN and Anchor Head modules sequentially to propose 3D Regions of Interest (ROI). During this process, the Anchor Head module predicts classification scores and 3D bounding boxes (3D Bboxes) for all objects using pre-defined anchor boxes at $0^\circ$ and $90^\circ$. Consecutively, the ROI Head module combines the predicted box information from the 1-stage with voxel and point features using voxel set abstraction (VSA), leveraging their rich context information for more refined 3D Bbox prediction. 

To achieve this, the VSA module generates the final feature $F_{final}$ by combining key points extracted from $V_{f}$, denoted as $V_{key}= \{V_{1k}, V_{2k}, V_{3k}, V_{4k}$\}, with raw points sampled using the furthest point-sampling (FPS) technique as $p_{key}$, and the interpolated 3D proposed ROI as BEV features, denoted as $BEV_{key}$. This combined feature $F_{final}$ is fed into ROI grid pooling and two MLP modules in a 2-stage method, ultimately predicting the confidence scores and refined 3D Bboxes. All these processes are supervised by the following loss function $L^l$:

\begin{equation}
L^l = L^{rpn}_{cls} + L^{rpn}_{reg} + L^{rcnn}_{iou} +L^{rcnn}_{reg},
\end{equation}\label{eq1}

\noindent Here, $L^{rpn}_{cls}$ represents the loss function for the classification branch in the 1-stage, which is the Focal Loss~\cite{bib31}. $L^{rcnn}_{iou}$ denotes the loss function for the IoU estimation branch, which predicts confidence scores in the 2-stage. The loss function for this IoU estimation branch is optimized using binary-cross entropy loss based on the IoU between 3D ROIs and their corresponding foreground GT boxes. $L^{rpn}_{reg}$ and $L^{rcnn}_{reg}$ refer to the box regression losses in the 1-stage and 2-stage, respectively, both calculated using the Smooth-L1 loss. After completing Burn-In Stage, the weights and structure of the pre-trained PV-RCNN are duplicated for multiple teachers and a single student network.

\noindent \textbf{MultipleTeachers Fine-tuning Stage}. 
In this stage, we fine-tune each teacher network, which has been comprehensively pre-trained on all classes information, to focus on specific category groups with similar aspect ratios and sizes. This fine-tuning process aims to generate more accurate pseudo-labels (see Fig.~\ref{fig3} (b)). During this process, each teacher network focuses on learning the characteristics (e.g., aspect ratio, size, and shape) of its assigned category group. In contrast, labels from other categories are treated as background. This training strategy helps to reduce influence between different category groups. By employing this method, each teacher network can guide the student network to learn using high-quality pseudo-labels optimized for its respective category. Consequently, these generated high-quality pseudo-labels significantly enhance the object detection performance of the student network.

\noindent \textbf{Mutual Learning Stage}. 
During this stage, we employ both a large volume of unlabeled data and a small subset of labeled data to apply mutual learning between multiple teachers network and a single student network. Throughout this process, multiple expert teacher networks work collaboratively to generate high-confidence pseudo-labels. The student network is then trained with these pseudo-labels to gradually improve its object detection accuracy. This process is detailed in Fig.~\ref{fig2}. Initially, the MPGen module applies weak augmentation techniques (such as random flipping, rotation, and scaling) to the unlabeled data. Each category-wise specialized teacher network then predicts pseudo-label candidates from the augmented data. These predicted candidates performs non-maximum suppression (NMS) to eliminate duplicate boxes. After that, the remaining candidate boxes are merged with those predicted by other teachers. The aggregated candidate set, $\bar{y}{candi}$, is then filtered using dynamic thresholds based on object confidence, classification, and IoU consistency, ensuring that only pseudo-labels with high positional accuracy and reliability, $\bar{y}{select}$, are selected. Thus, the final selected pseudo-labels generated by the MPGen module, $\mathit{\bar{y}_{select}}$, are subsequently used to train the student network.

To further enhance the training efficiency of the student network, we additionally introduce a novel augmentation method named PieAug. PieAug first divides the point clouds into fixed regions and then compensates areas with insufficient foreground points by supplementing them with points from other dense regions. This improves the quality of the training data. A detailed explanation of this technique is provided in Sec.~\ref{sec3.4}. Through this process, the student network learns from both the augmented unlabeled data \{$\bar{p}$, $\bar{y}$\} and the labeled data \{$\hat{p}$, $\hat{y}$\}. Consequently, the student network predicts classification scores $s_{cls}$ and bounding boxes $b_{cls}$ for objects from the input point clouds using a 2-stage approach. These results are then used to train the student network with the total loss function $L^{total}$, which is computed as follows:

\begin{equation}
L^{total} = L^l + \lambda \times L^u,
\end{equation}\label{eq2}

\noindent where $L^l$ represents the loss for labeled data, and $L^u$ represents the loss for unlabeled data. $\lambda$ is the weighting factor for the loss from unlabeled data, which is set to 1. Based on this loss function, the weights of the student network are updated through back-propagation. Successively, the weights of the category-wise specialized multiple teachers network are updated by utilizing the category-wise EMA (C-EMA) strategy:

\begin{equation}
 \theta^{t} = \alpha \cdot  \theta^{t-1} + (1-\alpha) \cdot \theta^{s}
\end{equation}\label{eq3}

\noindent where $\alpha$ represents the EMA momentum. This parameter is decay weight of between the weight of the previous iterations of the multiple teachers network $\theta^{t-1}$ and the current weights of the student network $\theta^{s}$, and it is used to calculate the current weights of multiple teachers network $\theta^{t}$. Here, the weights of the networks consist of the Backbone, Neck, Anchor Head, and RCNN Head modules.

Ultimately, the C-EMA process at the Anchor Head involves updating the weights of multiple teachers network using the weights of student network, which have different dimensions for each category. Due to the varying weight dimensions between the networks, we employ new $Adjust$ function to split the weights of student network, $\theta^{s}_{anc}$, by category. This function is performed as follows:

\begin{equation}
\theta^{s'}_{anc} = Adjust(\theta^s_{anc}), \; \text{where} \; \theta^{s'}_{anc} \in \mathrm{R}^{C_{out} \times C_{in} \times p \times p}
\end{equation}\label{eq4}

\noindent where $\theta^{s'}_{anc}$ represents the weights for the Anchor Head in the student network, after applying the $Adjust$. In addition, $C_{in}$ denotes the number of input channels, while $C_{out}$ represents the number of output channels. The kernel size is defined as $p$, which is set to 3. The output channels are calculated by multiplying three elements, such as $\beta$, $dim$, and $out$. Here, $\beta$ refers to the pre-defined orientations of the anchor, specifically $0^\circ$ and $90^\circ$. The dimensions of the anchor boxes are denoted as $dim$, which includes the center position ($c_x$, $c_y$, $c_z$), size ($l$, $w$, $h$), and heading ($yaw$). $out$ means the class-wise indices. In this paper, $Adjust$ function is used to split the range of class-wise indices $out = \{1, 2, \ldots, z\}$ from the student network, during the C-EMA process. The indices divided by $Adjust$ are assigned to three categories: the vehicle category from 1 to $m-1$, the pedestrian category from $m$ to $n-1$, and the cyclist category from $n$ to $z$.

By utilizing C-EMA, our MultipleTeachers network progressively updates its weights by incorporating feedback from the student parameters, thereby focusing more on the detailed information of each category. Consequently, our teacher network is gradually guided to generate higher quality pseudo-labels by reflecting the distinctive characteristics of each category (see Fig.~\ref{fig6}). These generated pseudo-labels are used in the training process of the student network, which, guided by these reliable pseudo-labels, becomes increasingly sophisticated. This collaborative guidance results in a significant improvement in object detection performance.

\begin{figure}[t]
\centering
\includegraphics[width=1.0\textwidth]{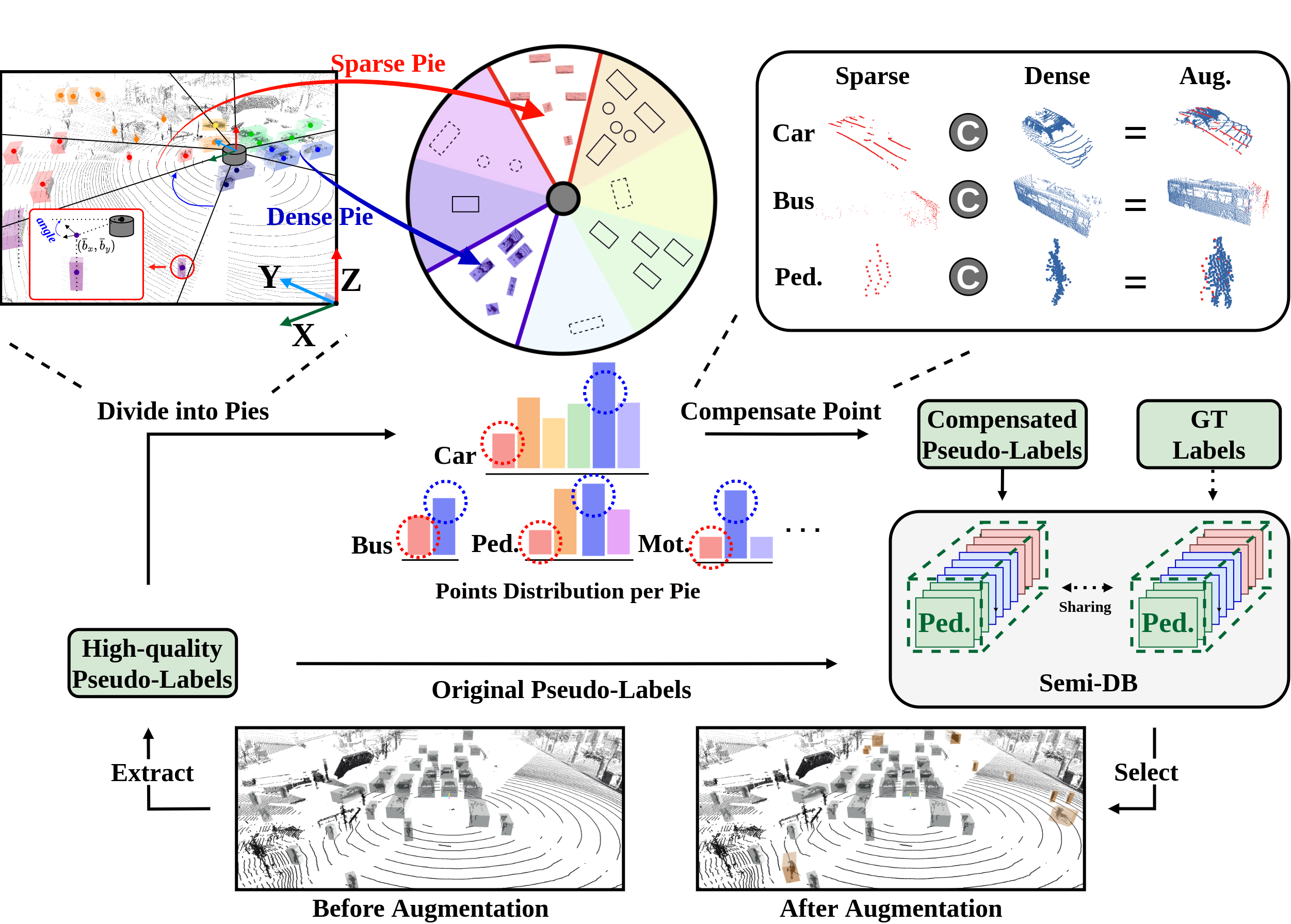}
\caption{\textbf{Illustration of the PieAug process.} Point clouds are divided into several pies of equal size, classified by the density of objects within each pie as either sparse or dense. The points from dense pies compensate for the lack of points in sparse pies. It is lead to the student network to learn from enriched features of point data. Note that, the gray boxes represent before augmentation, while the brown boxes indicate boxes newly added through augmentation}\label{fig4}
\end{figure}

\begin{algorithm}
\caption{Pie-based point compensating augmentation}\label{algo1}
\begin{algorithmic}[1]
\Require $ \text{Point clouds: } \; \bar{p}, \; \text{pseudo-labels: } \; \bar{y}_{select} = \{ \bar{c}_i, \bar{s}_i, \bar{b}_i \}^{N}_{i=1} $
\Ensure \text{Augmented point clouds: } $\bar{p}_{final}$,\; \text{augmented pseudo-labels: } $\bar{y}_{final}$
\State \textbf{Initialization}

\State  Augmented foreground points $\bar{p}^{obj}_{pie} \leftarrow \phi$, $\bar{p}_{final} \leftarrow \phi, \bar{y}_{final} \leftarrow \phi$

\For { each  $\{ \bar{c}_i, \bar{s}_i, \bar{b}_i \} \in \bar{y}_{select}$}
\State Extract foreground points $\bar{p}^{obj}_{i}$ within $\bar{b}_{i}$.
\State  $pie_k$ = \{$\bar{p}^{obj}_{i}, \bar{b}_i \}$, where $k$ is followed to Eq.(5)
\EndFor

\State Sort $pie$ in descending order by points density according to Eq.(6)
\While{${pie} \neq \phi$}
    \State $pie_{d} \leftarrow pie[1]$ $\qquad\qquad\quad \slash \star$ Extract the most dense pie $ \star \slash$
    \State $pie_{s} \leftarrow pie[end]$ $\qquad\qquad \slash \star$ Extract the most sparse pie $ \star \slash$
    \State $M \leftarrow$ min($N^{obj}_{d}$, $N^{obj}_{s}$) $\;\;\; \slash \star M$ is minimum number of objects $\star\slash$
    \For {$j= 1, 2, \ldots,  M$}  
        \State $ \bar{p}^{obj}_{d'(j)} \leftarrow  F_{trans}(F_{rot}(F_{sca}(\bar{p}^{obj}_{d(j)} )))$, where $\bar{p}^{obj}_{d(j)} \in pie_{d}$
        \State $\bar{p}^{obj}_{aug(j)} \leftarrow Concat(\bar{p}^{obj}_{s(j)},\bar{p}^{obj}_{d'(j)})$, where $\bar{p}^{obj}_{s(j)} \in pie_{s}$
        \State $\bar{p}^{obj}_{pie} \leftarrow  \bar{p}^{obj}_{pie} \cup \bar{p}^{obj}_{aug(j)} \cup \bar{p}^{obj}_{d(j)}$
    \EndFor
    \State $pie \leftarrow  pie \; \backslash  \; pie_d$    $\qquad\qquad \slash \star$ Remove  $pie_d, pie_s \star \slash$
    \State $pie \leftarrow  pie \; \backslash  \; pie_s$    
\EndWhile
\State Generate semi-DB from \{$\bar{y}_{select}$, $\bar{p}^{obj}_{pie}$\}
\For {$t= 1, 2, \ldots, T $} $\qquad\quad \slash \star T$ is number of total frames $\star \slash$
    \State Select \{$\bar{y}_{c}$, $\bar{p}_{c}$\} from semi-DB
    \State $\bar{y}_{final} \leftarrow Concat(\bar{y}_t, \bar{y}_{c})$
    \State $\bar{p}_{final} \leftarrow Concat(\bar{p}_t, \bar{p}_{c})$  
\EndFor
\end{algorithmic}
\end{algorithm}

\subsection{Pie-based Point Compensating Augmentation}\label{sec3.4}
The primary goal of this research is to enable the student network to achieve state-of-the-art object detection performance by leveraging robust feature information obtained from diverse scenes. To achieve this, it is essential to employ strong augmentation techniques~\cite{bib15, bib21, bib28, bib29} that significantly transform point clouds and increase the diversity of foreground objects. However, existing 3D SSOD approaches heavily rely on weak augmentations, which are quite limited in effectiveness due to the lack of color information and the inherent sparsity of LiDAR point clouds. To address these challenges, we propose a novel strong augmentation strategy called PieAug, which is meticulously designed to address the sparsity of 3D point clouds.

To this end, it focuses on enhancing object diversity and accuracy by analyzing the point distribution in foreground objects. In sparse regions where objects are difficult to detect, the object is compensated by adding points from dense regions. This technique improves the accurate representation of object information and increases the variety of objects through point cloud transformation. The approach involves partitioning the point clouds into pie-shaped sectors with equal angles, to minimize the issues caused by the imbalance in foreground point distribution. As illustrated in Fig.~\ref{fig4}, firstly, we divide the entire point clouds into several pie-shaped spatial regions with a regular angle $deg$, around the LiDAR center. These regions are then classified into two groups: sparse pies and dense pies, according to the density of foreground points. The objects assigned in sparse pies are compensated with additional points from objects in dense pies. This design enables the student network to learn from more enriched and distinctive point information. Therefore, this augmentation strategy plays a crucial role in enabling the student network to successfully perform robust generalization and object detection capabilities.

As described in Algorithm~\ref{algo1}, PieAug takes the unlabeled point cloud $\bar{p}$ and the high-quality pseudo-labels $\bar{y}_{select}$ produced by multiple teachers, as initial inputs. Here, $\bar{y}_{select}$ consists of $N$ pseudo-boxes $\bar{b}$, their respective classes $\bar{c}$, and confidence scores $\bar{s}$. This process includes dividing the pseudo-boxes $\bar{b}$ within a 360$^\circ$ range around the LiDAR into 360/$deg$ pie-shaped sectors. After partitioning sectors, foreground points $\bar{p}^{obj}$ are extracted from $\bar{b}$ (refer to line 4 in Algorithm~\ref{algo1}) and assigned to the corresponding pie sector $k$ using the following equation:

\begin{equation}
k =  \left\lfloor \frac{(\text{atan2}(\bar{b}_x, \bar{b}_y)+180^\circ) \times 180^\circ}{\pi \times deg} \right\rfloor,  \; \text{where} \; \bar{b} \in \bar{y}_{select},  \; 0 \leq k < \frac{360^\circ}{deg},
\end{equation}\label{eq5}

\noindent where we first calculate the horizontal angle of each pseudo-box $\bar{b}$ by leveraging its center point ($\bar{b}_{x}$, $\bar{b}_{y}$). The radian values of horizontal angle are then converted to degree values, and the angle is divided by the pre-defined $deg$ value to compute $k$, where $k$ means pie ID and it is used to assign each $\bar{p}^{obj}$ and $\bar{b}$ to the corresponding $pie_{k}$ in line 5 of Algorithm~\ref{algo1}.

This technique is particularly focused on by supplementing the sparse pie-shaped sectors $pie_s$, where objects are either far from the LiDAR sensor or occluded by other objects, which have fewer points. By supplementing the points in these sparse pies with those from objects in dense pies $pie_d$ with relatively lots of points, we enable to achieve a data augmentation effect. Experiments reveal that dense pies typically contain large objects such as buses or trucks, or objects that are close to the LiDAR sensor. This is because the LiDAR sensor generally captures more points on the surfaces of objects that are facing it. This pattern is especially observed when the objects are closer or larger. To address the issue of point density bias from on these objects, we normalize the density of each pie based on the number of objects and the number of points per object. This class-wise normalization process is defined as follows:

\begin{equation}
Norm(density) = \frac{1}{N^{obj}} \times \sum_{i}^{N^{obj}} (\bar{p}^{obj}_{i}), \; \text{where} \; \bar{p}^{obj}_{i} \in pie, 
\end{equation}\label{eq6}

\noindent where $N^{obj}$ represents the number of objects per pie, and $\bar{p}^{obj}$ denotes the number of foreground points of those objects. After normalization process, the pies are sorted according to the values to classify high-density pies (${pie}_d$) and low-density pies (${pie}_s$) (Algorithm~\ref{algo1}, lines 7 to 10).

The points from objects in $pie_d$, defined as $\bar{p}^{obj}_d$. They are adjusted for size using global scaling $F_{sca}(\cdot)$. After that, these adjusted points are rotated using $F_{rot}(\cdot)$ and translated using $F_{trans}(\cdot)$ to align with the positions of the target objects in the sparse pie, $pie_s$ (line 13). After that, these transformed points $\bar{p}^{obj}_{d'}$ are combined with the points $\bar{p}^{obj}_s$ of target objects. In addition, the augmented points $\bar{p}^{obj}_{pie}$ is updated by merging the previous combined points $\bar{p}^{obj}_{aug}$ and the original points $\bar{p}^{obj}_d$ in dense pie, by iterating for the number of objects. 

The augmented points $\bar{p}^{obj}_{pie}$ and pseudo-labels $\bar{y}_{select}$ are subsequently stored in the semi-DB. They are then utilized during the mutual learning stage of MultipleTeachers framework. This data augmentation process significantly increases object diversity by introducing randomly augmented 3D pseudo-labels, $\bar{y}_{c}$, and points, $\bar{p}_{c}$, into background areas where no objects are initially present. While this study has focused on explaining the overall process of PieAug for generating pseudo-labels, it is important to note that this technique can be flexibly applied to both labeled and unlabeled data.

\begin{table}[h]
\caption{Specification of labeling guidelines for ambiguous classes of the proposed LiO dataset.}\label{tab1}%
\setlength{\tabcolsep}{11.0pt}
\renewcommand{\arraystretch}{1.4}
\begin{tabular}{l|l}
\hline
Class                                                                     & Object                                                                     \\ \hline
\multirow{3}{*}{Car}                                                      & Trailer, Police Car                                                        \\
                                                                          & SUV, LCV, Van                                                              \\
                                                                          & Vehicle being towed, Vehicle trailer                                       \\ \hline
\multirow{2}{*}{Truck}                                                    & Camping car                                                                \\
                                                                          & Trike truck, Box truck, Tank car                                           \\ \hline
Bus                                                                       & All types of buses (medium, large)                                         \\ \hline
\multirow{2}{*}{Motorcycle}                                               & Parked or moving motorcycle                                                \\
                                                                          & Electric wheelchair                                                        \\ \hline
Bicycle                                                                   & Person carrying bicycle                                                    \\ \hline
\multirow{2}{*}{Pedestrian}                                               & Person carrying children, Stroller person                                  \\
                                                                          & Person with electric kickboard or wheeled device                           \\ \hline
\multirow{3}{*}{\begin{tabular}[c]{@{}l@{}}Other\\ vehicles\end{tabular}} & Forklift, Mixer Truck, Construction vehicle                                \\
                                                                          & Fire engine, Ambulance                                                     \\
                                                                          & Auto transporter, Sweeper vehicle, Ladder truck, Cultivator, Tractor truck \\ \hline
\multirow{2}{*}{\begin{tabular}[c]{@{}l@{}}No\\ labeling\end{tabular}}    & Bike racks                                                                 \\
                                                                          & Parked electric scooters                                                   \\ \hline
\end{tabular}
\end{table}

\subsection{LiO Dataset: 3D LiDAR Open Dataset for Autonomous driving}\label{sec3.5}
As a research group focused on autonomous driving perception technology, we determine that building a dataset representing domestic urban environments is essential before developing advanced deep learning-based perception technologies. Consequently, we construct the LiO dataset, a 3D LiDAR open dataset for autonomous driving, which reflects a variety of objects and driving environments in urban settings. In this section, we provide a detailed description of the process we used to build our dataset and conduct a comparative analysis with other public datasets.

\noindent \textbf{Data Collection and Filtering}. 
As the first step in constructing our dataset, we equip a vehicle with sensors to collect data. This vehicle is fitted with a high-resolution 128-channel LiDAR (RS Ruby) and six cameras to gather data in various road environments. To minimize data bias, we collect extensive data from diverse road environments such as complex urban areas, intersections, residential zones, and highways. Prior to data collection, we carefully select the target classes to be annotated, considering class balance. The primary labeling targets include large objects such as cars, buses, trucks, and other vehicles, as well as smaller objects like pedestrians, bicycles, and motorcycles. The goal is to annotate approximately 21,000 frames as labeled datasets, with a plan to ensure that at least 20\% of the dataset consist of pedestrians and at least 4\% include bicycles and motorcycles.

To ensure the acquisition of high-quality data across diverse scenarios, we focus on establishing a robust dataset through systematic and detailed data refinement guidelines, implemented in a 2-stage process. In the first refinement stage, we identify and remove anomalous data caused by technical issues, such as sensor malfunctions or failure. The second refinement stage involves a thorough review to eliminate frames that degrade data quality, such as those without objects, static frames with no changes, or frames from areas susceptible to localization errors, like tunnels. This meticulous process ensures that only the data is selected for labeling. Finally, to enhance the diversity and comprehensiveness of the scenarios, we down-sample the acquired 10Hz LiDAR data to 2Hz.

\noindent \textbf{Data Annotation and Quality Management}.
The 3D data annotation process described in this paper starts with generating initial pseudo-labeling results using a pre-trained 3D detector on a public dataset \cite{bib41}. These pseudo-labeled data are then meticulously reviewed by annotators to achieve more accurate GT labels. The following labeling correction steps are conducted to enhance labeling accuracy. First, the annotators accurately identify and correct any FP and FN. This involves generating new bounding boxes for missed objects or modifying existing bounding boxes. In cases where class definitions are ambiguous, annotators refer to the labeling guidelines in Table~\ref{tab1} to determine the correct class.

Specifically, for objects that are occluded or distant, making accurate recognition challenging, annotators reference camera images or analyze consecutive frames to more precisely adjust position and size of the object. After the annotation process is completed, a professional reviewer conducts a thorough quality check and makes iterative corrections as needed to maximize data accuracy. The finalized data are then used for training deep-learning models, and through a continuous cycle of user feedback and model training, the data quality is progressively enhanced.

\begin{table}[h]
\caption{Comparison with the LiO dataset and other autonomous driving datasets}\label{tab2}%
\setlength{\tabcolsep}{7.5pt}
\renewcommand{\arraystretch}{1.4}
\begin{tabular}{l|cccccc}
\hline
Datasets                         & \#Scene     & \#Frame       & FOV                    & Channels     & \#Classes  & \#3DBox        \\ \hline
KITTI \cite{bib4}              & -           & 15 K          & -                      & 64           & 3          & 80 K           \\
H3D \cite{bib33}               & 160         & 27 K          & $360^{\circ}$          & 64           & 8          & 1.1 M          \\
ONCE (only Label) \cite{bib32} & 581         & 10 K          & $360^{\circ}$          & 40           & 5          & 417 K          \\
Waymo \cite{bib3}              & 1,150       & 230 K         & $360^{\circ}$          & -            & 4          & 12 M           \\
\textbf{LiO Label Set}           & \textbf{35} & \textbf{21 K} & \textbf{$360^{\circ}$} & \textbf{128} & \textbf{7} & \textbf{755 K} \\ \hline
LiO Unlabel Set                  & 31          & 96 K          & $360^{\circ}$          & 128          & -          & -              \\ \hline
\end{tabular}
\end{table}

\begin{figure}[t]
\centering
\includegraphics[width=1.0\textwidth]{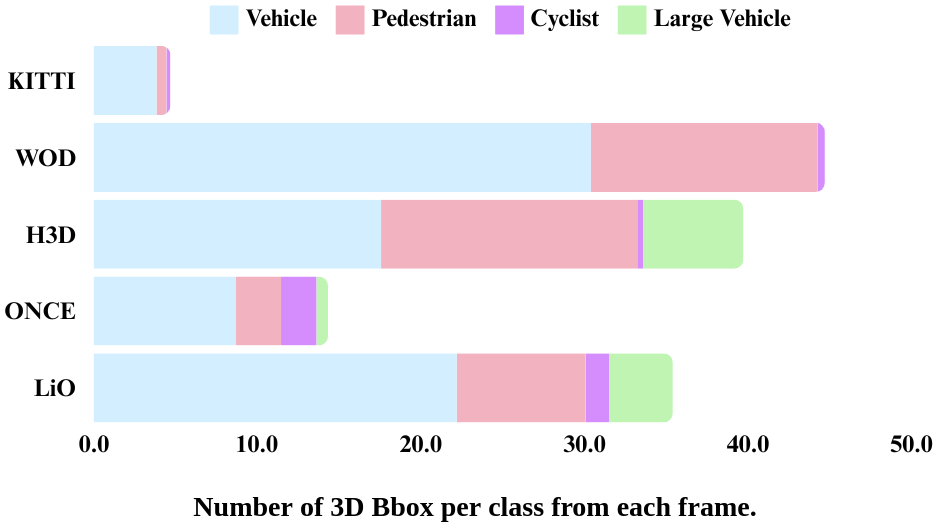}
\caption{\textbf{Number of annotated 3D bounding boxes per category for each frame.} Each frame contains 45 boxes for WOD, 38 boxes for H3D, 15 boxes for the ONCE, 5 boxes for KITTI, and 36 boxes for our LiO dataset, respectively.}\label{fig5}
\end{figure}

\noindent \textbf{Comparison with Other Autonomous Driving Datasets}.
As shown in Table~\ref{tab2}, the LiO dataset consists of 35 scenarios. Each scenario includes point cloud data and GT labels for approximately 600 frames collected using a high-resolution 128-channel LiDAR installed on a vehicle. This high-resolution sensor provides significantly richer point cloud information compared to LiDAR-based datasets with fewer channels \cite{bib5,bib32}. This richness allows annotators to generate 3D GT bounding boxes more easily and accurately, greatly enhancing the overall quality of the dataset.

The LiO dataset includes extensive scenarios such as complex urban environments, residential areas, and intersections. Unlike the KITTI dataset~\cite{bib4}, which has a limited field of view (FOV), our dataset features a 360-degree FOV, thereby boosting the diversity of the training data. Additionally, the LiO dataset is meticulously annotated for seven classes, more detailed than those in WOD~\cite{bib3}. On average, each frame contains 35.8 objects, which is one of the highest counts following WOD and H3D (see Fig.~\ref{fig5}). These statistical results indicate that the LiO dataset is collected in complex road environments, with meticulous annotation process. This highlights the value of the LiO dataset as a baseline dataset for advancing 3D object detection method, compared to public datasets.

In this study, 35 scenarios of labeled datasets, comprising approximately 21K frames, are divided into 27 scenarios for training sets and 8 scenarios for test sets. These training sets include 16K frames, while the test sets contain 5K frames. In addition, to train the 3D SSOD models, the unlabeled datasets consist of 96K unlabeled frames. This setup enables us to train various detectors and evaluate their performance. For performance evaluation, we calculate the IoU-based 3D Average Precision (AP) for seven classes: car, bus, truck, other vehicles, pedestrian, motorcycle, and bicycle. In addition, we group all classes into three categories such as, car, pedestrian, and cyclist on their similarities. We then compute the 3D Mean Average Precision (mAP) for these grouped categories to evaluate the overall performance of the detectors. Specifically, during performance evaluation, the IoU thresholds are set to 0.7 for car, bus, and truck, and 0.5 for other vehicles, pedestrian, motorcycle, and bicycle.

\section{Experiments}\label{sec4}
To verify the superiority of the proposed MultipleTeachers framework, we conduct comparative analyses with existing state-of-the-art models on the KITTI \cite{bib3}, WOD \cite{bib4}, and LiO datasets.

\begin{table}[h]
\caption{Experimental comparison with recent state-of-the-art methods on KITTI validation set. This evaluation results are calculated based by average precision (AP) with IoU threshold 0.7 for car, and 0.5 for pedestrian and cyclist at 40 recall positions}\label{tab4}%
\setlength{\tabcolsep}{2.2pt}
\renewcommand{\arraystretch}{1.4}
\begin{tabular}{l|c|cccc|cccc|cccc}
\hline
\multicolumn{1}{c|}{\multirow{2}{*}{Methods}} & \multirow{2}{*}{SSL} & \multicolumn{4}{c|}{1\%}                                                                                            & \multicolumn{4}{c|}{2\%}                                                                                            & \multicolumn{4}{c}{15\%$\footnotemark[1]$}                                                                         \\
\multicolumn{1}{c|}{}                         &                      & \multicolumn{1}{l}{mAP}            & \multicolumn{1}{l}{Car} & \multicolumn{1}{l}{Ped.} & \multicolumn{1}{l|}{Cyc.} & \multicolumn{1}{l}{mAP}            & \multicolumn{1}{l}{Car} & \multicolumn{1}{l}{Ped.} & \multicolumn{1}{l|}{Cyc.} & \multicolumn{1}{l}{mAP}            & \multicolumn{1}{l}{Car} & \multicolumn{1}{l}{Ped.} & \multicolumn{1}{l}{Cyc.} \\ \hline
PV-RCNN \cite{bib23}                        &                      & \multicolumn{1}{c|}{43.5}          & 73.5                    & 28.7                     & 28.4                      & \multicolumn{1}{c|}{54.3}          & 76.6                    & 40.8                     & 45.5                      & \multicolumn{1}{c|}{67.9}          & 82.3                    & 55.1                     & 66.3                     \\ \hline
3DIoUMatch \cite{bib11}                     & \checkmark         & \multicolumn{1}{c|}{48.0}          & 76.0                    & 31.7                     & 36.4                      & \multicolumn{1}{c|}{61.0}          & 78.7                    & 48.2                     & 56.2                      & \multicolumn{1}{c|}{68.5}          & 81.5                    & 47.8                     & 68.5                     \\
DDS3D \cite{bib13}                          & \checkmark         & \multicolumn{1}{c|}{49.8}          & 76.0                    & 34.8                     & 38.5                      & \multicolumn{1}{c|}{60.7}          & 78.9                    & 49.4                     & 53.9                      & \multicolumn{1}{c|}{64.5}          & 81.5                    & 43.2                     & 68.6                     \\
Reliable Student \cite{bib14}               & \checkmark         & \multicolumn{1}{c|}{51.7}          & 77.0                    & 41.9                     & 36.4                      & \multicolumn{1}{c|}{63.8}          & 79.5                    & 53.0                     & 59.0                      & \multicolumn{1}{c|}{-}             & -                       & -                        & -                        \\
DetMatch \cite{bib12}                       & \checkmark         & \multicolumn{1}{c|}{59.0}          & 77.5                    & \textbf{57.3}            & 42.3                      & \multicolumn{1}{c|}{65.6}          & 78.2                    & 54.1                     & 65.6                      & \multicolumn{1}{c|}{-}             & -                       & -                        & -                        \\
HSSDA \cite{bib15}                          & \checkmark         & \multicolumn{1}{c|}{59.5}          & \textbf{80.9}           & 51.9                     & 45.7                      & \multicolumn{1}{c|}{68.6}          & \textbf{81.9}           & 58.2                     & 65.8                      & \multicolumn{1}{c|}{71.1}          & \textbf{84.1}           & 57.6                     & 71.5                     \\ \hline
\textbf{MultipleTeachers}                              & \checkmark         & \multicolumn{1}{c|}{\textbf{64.6}} & 80.2                    & 55.8                     & \textbf{57.7}             & \multicolumn{1}{c|}{\textbf{70.5}} & 81.1                    & \textbf{59.3}            & \textbf{71.2}             & \multicolumn{1}{c|}{\textbf{71.7}} & 82.3                    & \textbf{58.4}            & \textbf{74.4}            \\ \hline
\end{tabular}
\footnotetext{Note that, `Ped.' is pedestrian, and `Cyc.' means cyclist. `SSL' indicates semi-supervised learning, which utilizes a small amount of labeled data along with a substantial amount of unlabeled data}
\footnotetext[1]{For the 15\% split, we re-implemented existing 3D SSOD models to facilitate a performance comparison with our MultipleTeachers framework. This ensures an equivalent split for comparison with our proposed LiO dataset}
\end{table}

\subsection{Implementation Details}\label{sec4.1}
The MultipleTeachers is trained through a structured several stage learning pipeline: Burn-In, MultipleTeachers Fine-tuning, and Mutual Learning stages. During the Burn-In, only labeled data are utilized to pre-train the baseline model. Specifically, in the MultipleTeachers Fine-tuning stage, to generate specialized teachers optimized for each class group, we re-sample each labeled dataset according to category before proceeding with additional fine-tuning. During the Mutual Learning stage, the student model is trained using a combination of labeled and unlabeled data. To fair comparisons across all experiments, the batch size for the MultipleTeachers and other re-implemented models is set to 2. Notably, our framework employs PV-RCNN \cite{bib23} as the baseline detector, aligning with other state-of-the-art models. The MultipleTeachers is trained for a total of 80 epochs on KITTI, and 30 epochs each on WOD and LiO. In addtion, the learning rate is set to 0.01, and optimization is conducted using AdamW on a server equipped with 8 GPUs.

\begin{table}[h]
\caption{Performance comparison of the MultipleTeachers with other state-of-the-art 3D SSOD methods on WOD validation set. Note that, all methods, employing SSL techniques, are trained with the 1\% labeled and 99\% unlabel data splits, and the PV-RCNN~\cite{bib23} is baseline model for a fair comparison. Evaluation metrics are categorized into Level 1 (L1) and Level 2 (L2)}\label{tab3}%
\setlength{\tabcolsep}{1.5pt}
\renewcommand{\arraystretch}{1.4}
\begin{tabular}{l|c|c|cccccc}
\hline
\multirow{2}{*}{Methods}  & \multirow{2}{*}{SSL} & mAP                & \multicolumn{2}{c}{Vehicle AP / APH}    & \multicolumn{2}{c}{Pedestrian AP / APH} & \multicolumn{2}{c}{Cyclist AP / APH}    \\
                          &                      & L1/L2              & L1                 & L2                 & L1                 & L2                 & L1                 & L2                 \\ \hline
PV-RCNN \cite{bib23}    &                      & 27.7/25.7          & 48.5/46.2          & 45.5/43.3          & 30.1/15.7          & 27.3/15.9          & 4.5/3.0            & 4.3/2.9            \\ \hline
DetMatch \cite{bib12}   & \checkmark         & -/-                & 52.2/51.1          & 48.1/47.2          & 39.5/18.9          & 35.8/17.1          & -/-                & -/-                \\
HSSDA \cite{bib15}      & \checkmark         & 41.9/37.0          & 56.4/53.8          & 49.7/47.3          & 40.1/20.9          & 33.5/17.5          & 29.1/\textbf{20.9} & 27.9/\textbf{20.0} \\ \hline
\textbf{MT} & \checkmark         & \textbf{46.3/41.1} & \textbf{59.8/57.7} & \textbf{52.0/50.2} & \textbf{44.0/22.7} & \textbf{37.5/19.3} & \textbf{35.2}/19.5 & \textbf{33.9}/18.7 \\ \hline
\end{tabular}
\footnotetext{Note that, `MT' is our proposed MultipleTeachers}
\end{table}

\begin{table}[h]
\caption{Performance comparison of SSL detectors on LiO test set between our models and strong competitor, HSSDA~\cite{bib15}. Both detectors are trained on 16K labeled (15\%) and 96K unlabeled data (85\%). To ensure a fair performance comparison, both models employ the pre-trained PV-RCNN~\cite{bib23} as the baseline detector}\label{tab5}%
\setlength{\tabcolsep}{11.5pt}
\renewcommand{\arraystretch}{1.4}
\begin{tabular}{l|l|c|c|ccc}
\hline
\multirow{2}{*}{Split} & \multirow{2}{*}{Methods}                 & \multirow{2}{*}{SSL} & \multirow{2}{*}{mAP} & \multicolumn{3}{c}{Category}         \\
                       &                                          &                      &                      & Car  & Pedestrain    & Cyclist       \\ \hline
\multirow{3}{*}{15\%}  & PV-RCNN\footnotemark[1] \cite{bib23}                       & 55.4                 & 69.3 & 55.7          & 41.7          \\
                       & HSSDA\footnotemark[1] \cite{bib15}   & \checkmark         & 56.7                 & 70.0 & 55.2          & 45.0          \\
                       & \textbf{MultipleTeachers}                & \checkmark         & \textbf{57.8}        & 69.7 & \textbf{57.1} & \textbf{46.6} \\ \hline
\end{tabular}
\footnotetext[1]{This result is re-implemented by ourselves}
\end{table}

The processing ranges for the point clouds of each model are specified as follows: For KITTI, the range on the $X$ and $Y$ axes is set to [0, -40, 70.4, 40], and on the $Z$ axis, it is set to [-3, 1]. For the WOD and LiO datasets, the detection range for the $X$ and $Y$ axes is [-75, -75, 75, 75], with the $Z$ axis set to [-3, 3] for WOD and [-5, 5] for LiO, respectively. In addition, to achieve state-of-the-art performance, our MultipleTeachers adopts advanced techniques such as shuffle data augmentation and dual-dynamic thresholding \cite{bib15}. Moreover, the parameter $deg$ for PieAug, utilized in training, is optimized to $45^{\circ}$. This technique is applied to the semi-DB, which is updated every 5 epochs.

\subsection{Comparison with State-of-the-art Methods}\label{sec4.2}
\noindent \textbf{KITTI Results.} 
In the KITTI dataset, we initially evaluate the performance of our MultipleTeachers against other state-of-the-art 3D SSOD models across three distinct labeled data splits: 1\%, 2\%, and 15\%. This experiment incorporates both supervised learning (SL) approach, which utilizes only a small percentage of labeled data, and a semi-supervised learning (SSL) approach, which additionally employs the remaining 99\%, 98\%, and 85\% of unlabeled data corresponding to the 1\%, 2\%, and 15\% splits, respectively. Following the methodology established by 3DIoUMatch \cite{bib11}, all experimental results are obtained by calculating the average mAP from three different samples. As shown in Table~\ref{tab4}, MultipleTeachers achieves 64.6 mAP, 70.5 mAP, and 71.7 mAP for label splits of 1\%, 2\%, and 15\%, respectively, significantly outperforming existing state-of-the-art methods. Notably, our method boosts the detection of small objects such as pedestrians and cyclists by achieving high accuracy. In the label splits of 2\% and 15\%, our approach surpasses the previous state-of-the-art detector HSSDA \cite{bib15} by a margin of 1.1 to 5.4 AP. These results demonstrate that, MultipleTeachers significantly enhances detection performance, especially for small objects, when compared to existing models.

\noindent \textbf{WOD Results.} 
In the 1\% labeled data split experiments on the WOD dataset, we observed significant performance improvements with our MultipleTeachers compared to DetMatch \cite{bib12} and HSSDA \cite{bib15}, particularly in the vehicle and pedestrian categories (refer to Table~\ref{tab3}). Specifically, our approach brings significant improvement, achieving an increase of 7.6 AP on L1 in vehicle class and 4.5 AP on L1 in pedestrian class compared to DetMatch. Moreover, it surpasses HSSDA by achieving 3.9 AP on L1 higher accuracy in vehicle class and 3.4 AP on L1 higher accuracy in pedestrian class. These results demonstrate that MultipleTeachers outperforms the current state-of-the-art methods in the field of 3DSSOD.

\noindent \textbf{LiO Results.} 
To further validate the effectiveness of our MultipleTeachers, we re-implement the previous state-of-art detector, HSSDA and compare its performance with our MultipleTeachers on the LiO test dataset. As shown in Table~\ref{tab5}, MultipleTeachers achieves 57.8 mAP, which surpasses the performance of the HSSDA by 1.1 mAP. This consistent improvement of detection accuracy on various datasets clearly validates the efficacy of our approach.

\subsection{Ablation Studies}\label{sec4.3}
In this section, we conduct extensive ablation experiments to validate our primary designs. Unless otherwise specified, all ablation studies are conducted using only 16K LiO labeled training dataset. We experiment by dividing this 16K labeled training dataset into labeled and unlabeled data splits at specified ratios. The unlabeled data is created by removing labels from the original 16K labeled dataset. For examples, the 15\% labeled data split means that we use 15\% labeled data of the 16K dataset and the remaining 85\% unlabeled data. The baseline detector used in these experiments is Voxel-RCNN \cite{bib22}, which is trained for 30 epochs with a batch size of 2.

\subsubsection{Evaluation on different fractions of LiO dataset}\label{sec4.3.1}
Table \ref{tab6} displays the comparative results between HSSDA and our MultipleTeachers framework across various labeled and unlabeled splits within the LiO dataset. In particular, our method is trained using 1\%, 2\%, and 15\% labeled data splits, respectively. In these experiments, our proposed model outperforms HSSDA by significant margins of 7.1 mAP, 7.0 mAP, and 1.3 mAP for the 1\%, 2\%, and 15\% labeled data splits, respectively.
 
\begin{table}[h]
\caption{Performance comparison on the LiO dataset. Experiments are conducted with different ratios of labeled data splits, such as 1\%, 2\%, and 15\%
}\label{tab6}%
\setlength{\tabcolsep}{11pt}
\renewcommand{\arraystretch}{1.4}
\begin{tabular}{c|l|c|c|ccc}
\hline
\multirow{2}{*}{Splits} & \multirow{2}{*}{Methods}                    & \multirow{2}{*}{SSL} & \multirow{2}{*}{mAP} & \multicolumn{3}{c}{Category}                  \\
                        &                                             &                      &                      & Car           & Pedestrian    & Cyclist       \\ \hline
\multirow{3}{*}{1\%}    & Voxel-RCNN\footnotemark[1] \cite{bib22} &                      & 15.3                 & 29.9          & 10.5          & 5.6           \\
                        & HSSDA$\footnotemark[1]$ \cite{bib15}      & \checkmark         & 30.7                 & 38.5          & 30.3          & 23.3          \\
                        & \textbf{MultipleTeachers}                   & \checkmark         & \textbf{37.8}        & \textbf{45.6} & \textbf{43.4} & \textbf{24.5} \\ \hline
\multirow{3}{*}{2\%}    & Voxel-RCNN$\footnotemark[1]$ \cite{bib22} &                      & 27.7                 & 40.2          & 27.5          & 15.3          \\
                        & HSSDA$\footnotemark[1]$ \cite{bib15}      & \checkmark         & 39.3                 & 51.9          & 38.0          & 28.0          \\
                        & \textbf{MultipleTeachers}                   & \checkmark         & \textbf{46.3}        & \textbf{56.1} & \textbf{45.6} & \textbf{37.0} \\ \hline
\multirow{3}{*}{15\%}   & Voxel-RCNN$\footnotemark[1]$ \cite{bib22} &                      & 51.2                 & 66.1          & 47.1          & 37.5          \\
                        & HSSDA$\footnotemark[1]$ \cite{bib15}      & \checkmark         & 53.7                 & 67.6          & 50.2          & 43.2          \\
                        & \textbf{MultipleTeachers}                   & \checkmark         & \textbf{55.0}        & \textbf{68.1} & \textbf{52.7} & \textbf{44.3} \\ \hline
\end{tabular}
\footnotetext[1]{This result is re-implemented by ourselves}
\end{table}

\begin{table}[h]
\caption{Ablation study on the effect of each component in MultipleTeachers using 15\% 
 labeled data split. The last row presents the results achieved by utilizing MPGen and PieAug, respectively}\label{tab7}%

\setlength{\tabcolsep}{10.5pt}
\renewcommand{\arraystretch}{1.4}
\begin{tabular}{l|cc|c|ccc}
\hline
\multirow{2}{*}{Methods} & \multicolumn{2}{c|}{Components} & \multirow{2}{*}{mAP} & \multicolumn{3}{c}{Category}                                                           \\
                         & MPGen          & PieAug         &                      & \multicolumn{1}{l}{Car} & \multicolumn{1}{l}{Pedestrian} & \multicolumn{1}{l}{Cyclist} \\ \hline
Baseline                 &                &                & 51.7                 & 67.6                    & 50.2                           & 37.4                        \\
MultipleTeachers         & \checkmark   &                & 53.8                 & 67.5                    & 52.4                           & 41.5                        \\
\textbf{MultipleTeachers}         & \checkmark   & \checkmark   & \textbf{55.0}        & \textbf{68.1}           & \textbf{52.7}                  & \textbf{44.3}               \\ \hline
\end{tabular}
\end{table}

\begin{table}[h]
\caption{Ablation study on the degree-wise PieAug on 1\% labeled data split}\label{tab8}%
\setlength{\tabcolsep}{22.5pt}
\renewcommand{\arraystretch}{1.4}
\begin{tabular}{c|c|ccc}
\hline
\multicolumn{1}{l|}{\multirow{2}{*}{Degree}} & \multicolumn{1}{l|}{\multirow{2}{*}{mAP}} & \multicolumn{3}{c}{Category}  \\
\multicolumn{1}{l|}{}                        & \multicolumn{1}{l|}{}                     & Car  & Pedestrian & Cyclist \\ \hline
$15^{\circ}$                                 & 53.0                                      & 66.5 & 49.1       & 43.5    \\ \hline
$30^{\circ}$                                 & 54.0                                      & 66.3 & 50.2       & 45.6    \\ \hline
$\textbf{45}^{\circ}$                                 & \textbf{54.2}                             & \textbf{67.0} & \textbf{50.3}       & 45.2    \\ \hline
$60^{\circ}$                                 & 53.0                                      & 66.5 & 49.1       & 43.5    \\ \hline
\end{tabular}
\footnotetext{Note that, MultipleTeachers are trained over 10 epochs for all comparative experiments regarding the degree}
\end{table}

\begin{table}[h]
\caption{Comparison of different strong augmentation. These models are trained using 1\% labeled data split in the KITTI and LiO datasets, respectively}\label{tab9}%
\setlength{\tabcolsep}{6.0pt}
\renewcommand{\arraystretch}{1.4}
\begin{tabular}{c|l|l|c|ccc}
\hline
\multirow{2}{*}{Datasets} & \multirow{2}{*}{Methods}                            & \multirow{2}{*}{Strong Augmentation} & \multirow{2}{*}{mAP} & \multicolumn{3}{c}{Category}                  \\
                          &                                                     &                                      &                      & Car           & Pedestrian    & Cyclist       \\ \hline
\multirow{4}{*}{KITTI}    & \multirow{4}{*}{MultipleTeachers$\footnotemark[1]$} & PolarMix~\cite{bib28}              & 36.4                 & 73.8          & 12.9          & 22.5          \\
                          &                                                     & Shuffle Data~\cite{bib15}          & 43.7                 & 76.2          & 20.7          & 34.3          \\
                          &                                                     & PseudoAugment~\cite{bib29}         & 62.2                 & 80.7          & 50.8          & 55.0          \\
                          &                                                     & \textbf{PieAug}                      & \textbf{64.6}        & \textbf{80.8} & \textbf{55.3} & \textbf{57.8} \\ \hline
\multirow{3}{*}{LiO}      & \multirow{3}{*}{MultipleTeachers$\footnotemark[2]$} & Shuffle Data~\cite{bib15}          & 35.9                 & 43.0          & 38.4          & 26.3          \\
                          &                                                     & PseudoAugment~\cite{bib29}         & 39.0                 & 46.1          & 42.9          & 28.0          \\
                          &                                                     & \textbf{PieAug}                      & \textbf{39.1}        & 45.1          & \textbf{43.4} & \textbf{28.9} \\ \hline
\end{tabular}
\footnotetext[1]{Note that, the model is based on PV-RCNN~\cite{bib23} and is trained with both labeled and unlabeled data for 80 epochs}
\footnotetext[2]{The model is based on Voxel-RCNN~\cite{bib22} and is trained with both labeled and unlabeled data for 30 epochs}
\end{table}

\begin{table}[h]
\caption{Experimental results for different detectors on 15\% labeled data split. For a fair comparison, all models employ SSL components from MultipleTeachers, including MPGen and PieAug}\label{tab10}%
\setlength{\tabcolsep}{11.0pt}
\renewcommand{\arraystretch}{1.4}
\begin{tabular}{l|l|c|ccc}
\hline
\multirow{2}{*}{Detectors}                                    & \multirow{2}{*}{SSL Methods} & \multirow{2}{*}{mAP} & \multicolumn{3}{c}{Category}                  \\
                                                              &                              &                      & Car           & Pedestrian    & Cyclist       \\ \hline
\multirow{3}{*}{PV-RCNN$\footnotemark[1]$ \cite{bib23}}     & Baseline                     & 47.1                 & 63.5          & 45.6          & 32.3          \\
                                                              & HSSDA                        & 49.1                 & 64.4          & 46.5          & 36.4          \\
                                                              & \textbf{MultipleTeachers}    & \textbf{50.5}        & \textbf{65.8} & \textbf{49.2} & 36.4          \\ \hline
\multirow{3}{*}{Voxel-RCNN$\footnotemark[1]$ \cite{bib22}}  & Baseline                     & 51.2                 & 66.1          & 47.1          & 37.5          \\
                                                              & HSSDA                        & 53.7                 & 67.6          & 50.2          & 43.2          \\
                                                              & \textbf{MultipleTeachers}    & \textbf{55.0}        & \textbf{68.1} & \textbf{52.7} & \textbf{44.3} \\ \hline
\multirow{3}{*}{Re-VoxelDet$\footnotemark[1]$ \cite{bib10}} & Baseline                     & 56.7                 & 65.4          & 58.1          & 46.7          \\
                                                              & HSSDA                        & 58.8                 & 65.7          & 59.7          & 51.1          \\
                                                              & \textbf{MultipleTeachers}    & \textbf{59.7}        & \textbf{66.6} & \textbf{60.3} & \textbf{52.2} \\ \hline
\end{tabular}
\footnotetext[1]{This result is re-implemented by ourselves}
\end{table}

\subsubsection{Effect of each component of MultipleTeachers}\label{sec4.3.2}
Table~\ref{tab7} provides a detailed analysis of the effectiveness of the two main techniques by integrating the MPGen module and PieAug strategy into the baseline model. Firstly, when replacing the pseudo-label generator (PGen) that uses a single teacher with our MPGen, we observe that the performance is boosted, with 2.2 AP in the pedestrian category and 4.1 AP in the cyclist category. This improvement indicates that high-quality pseudo-labels generated by multiple professional teachers provide crucial information for the training of the student network, thereby enhancing performance. Furthermore, when both the MPGen module and the PieAug strategy are applied together, additional performance improvements of 0.6 AP, 0.3 AP, and 2.8 AP are observed in the car, pedestrian, and cyclist categories, respectively. These experimental results clearly shows that the proposed techniques consistently improve the accuracy of the detector. 

\subsubsection{Analysis of degree-wise PieAug}\label{sec4.3.3}
In this section, we analyze the impact of various degree values in the PieAug technique on model performance. According to the experimental results presented in Table~\ref{tab8}, setting the degree to $45^{\circ}$ gains the best performance with 54.2 mAP. This is likely because a too-small degree results in insufficient object diversity within the pie, leading to insufficient point compensation. Conversely, a too-large degree reduces the total number of pies, making it challenging to distinguish between sparse and dense pies. Therefore, setting an optimal degree value is crucial for achieving the best performance with the PieAug technique. These findings indicate that careful parameter tuning is essential in enhancing the efficacy of our approach.

\begin{figure}[t]
\centering
\includegraphics[width=1.0\textwidth]{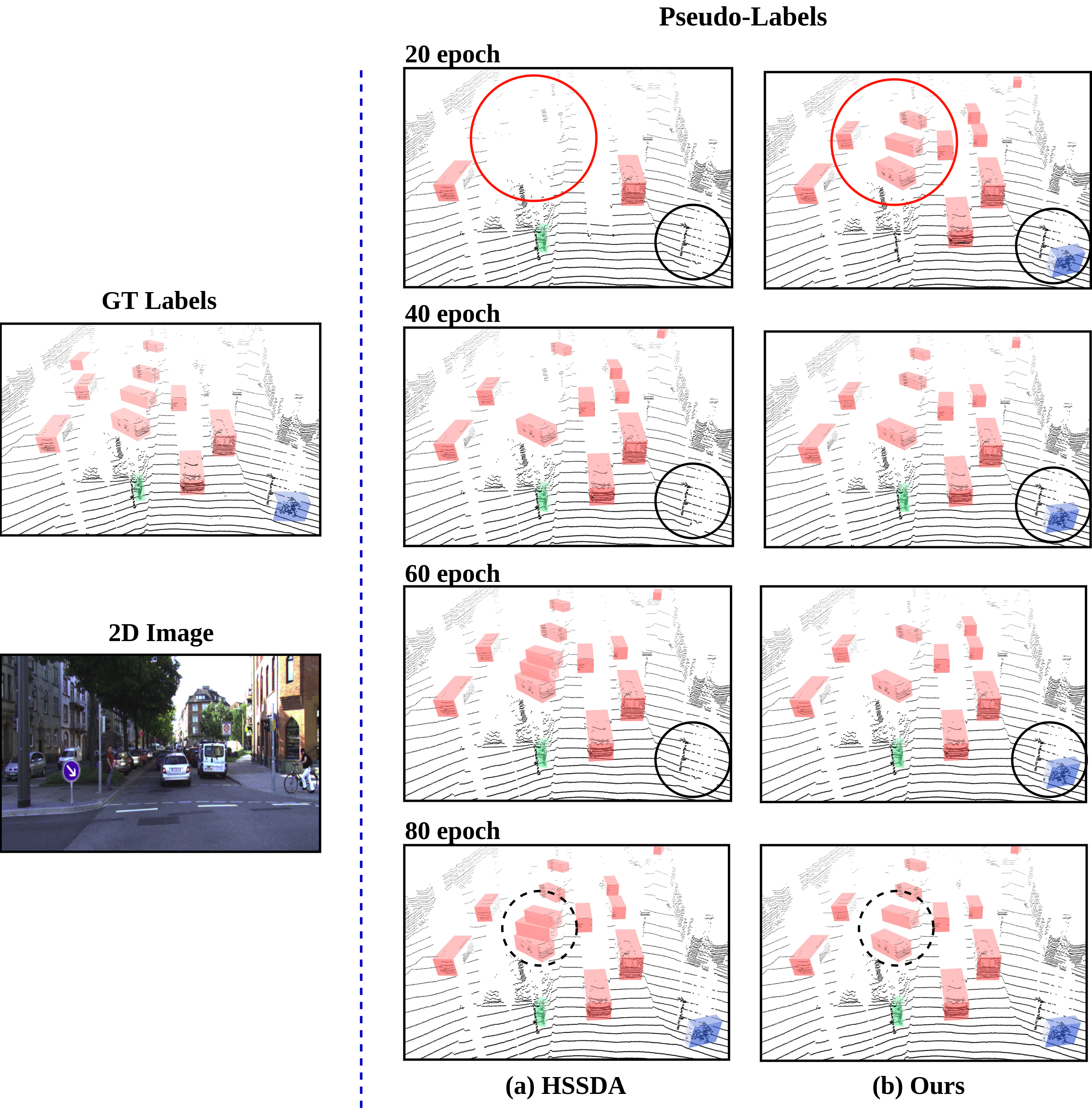}
\caption{Qualitative comparison of our MultipleTeachers with previous approach on the KITTI 1\% labeled data split. This comparison highlights the superiority of our proposed method by evaluating the quality of pseudo-labels generated from the teacher network every 20 epochs}\label{fig6}
\end{figure}

\subsubsection{Effect of PieAug}\label{sec4.3.4}
To verify the advantages of PieAug, we conduct comparative analyses to evaluate its performance against various strong augmentation strategy. As shown in Table~\ref{tab9}, we apply PolarMix \cite{bib28}, Shuffle Data Aug \cite{bib15}, Pseudo Boxes \cite{bib29}, and our PieAug strategy separately within the MultipleTeachers framework, and evaluate their impacts on object detection accuracy.

\begin{figure}[t]
\centering
\includegraphics[width=0.9\textwidth]{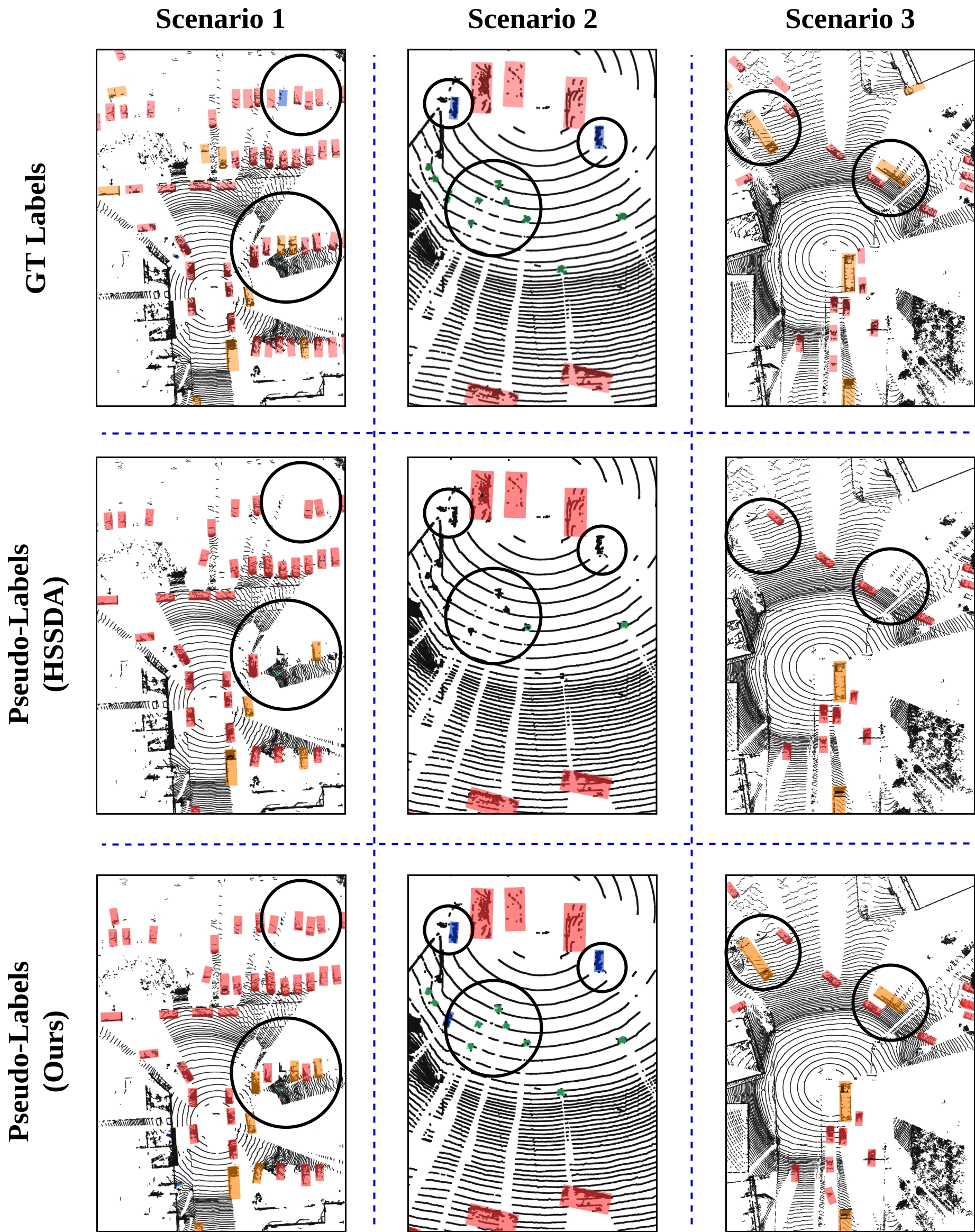}
\caption{Qualitative comparison of pseudo-labels on the LiO 1\% labeled data split. The top column displays ground truth boxes, the middle column presents pseudo-labels predicted by HSSDA, and the last column shows pseudo-labels predicted by our MultipleTeachers. This comparison highlights the enhanced accuracy and reliability of pseudo-labels generated by our approach}\label{fig7}
\end{figure}

As a result, PieAug achieves 64.6 mAP on the KITTI dataset and 39.0 mAP on the LiO dataset, surpassing previous strong augmentation methods. Notably, both our PieAug and PseudoAugment strategies outperform other existing strong augmentation methods due to their innovative use of copy-paste techniques. These techniques effectively increase the diversity of objects, which can markedly improve the accuracy of 3D SSOD techniques that primarily handle unlabeled data. Moreover, compared to PseudoAugment, our PieAug shows substantial improvements in the pedestrian category by 4.5 AP and in the cyclist category by 2.8 AP on the KITTI dataset. These dramatic enhancements reveal that, our strategy significantly boosts the detection performance of small objects, often have insufficient point clouds. This is because, our PieAug directly duplicates points from dense pie-shaped sectors to sparse pie-shaped sectors. This approach not only prevents the loss of geometric information but also provides higher quality training data to the student network.

\subsubsection{Robustness on different detectors}\label{sec4.3.5}
We evaluate the generalization ability and detection performance of our MultipleTeachers framework by comparing its SSL strategy against the previous state-of-the-art HSSDA on various detectors. As shown in Table~\ref{tab10}, when using detectors such as PV-RCNN, Voxel-RCNN, and Re-VoxelDet \cite{bib10}, our MultipleTeachers consistently achieves higher accuracy than HSSDA. These results demonstrate that our approach offers superior generalization capabilities and can be effectively applied to a wide range of detectors, highlighting its effectiveness and robustness in the field of 3D SSOD.

\subsection{Visualization Results}\label{sec4.4}
In this subsection, we qualitatively evaluate the quality of pseudo-labels generated by the teacher network, which critically impacts the detection performance of the student network. As shown in Fig.~\ref{fig6}, we visualize and compare pseudo-labels generated at 20, 60, and 80 epochs using 1\% labeled data split on the KITTI dataset. The HSSDA model notably fails to generate accurate pseudo-labels for the cyclist class even after several epochs, and its vehicle class predictions remain inaccurate even at 80 epochs. In contrast, MultipleTeachers framework consistently produces accurate pseudo-labels for both cyclist and vehicle objects as overall training progresses. This indicates that the progressively accurate weight updates of the teacher network through C-EMA significantly enhance the quality of pseudo-labels, ultimately improving the detection performance.

Fig.~\ref{fig7} presents the experimental results on the LiO dataset, comparing the quality of pseudo-labels generated by the HSSDA and MultipleTeachers methods through various scenarios. The first row shows the GT boxes, the second row displays the predictions from the HSSDA Teacher, and the last row presents the predictions from the MultipleTeachers method. Particularly, in the areas highlighted by circle lines, our approach demonstrates the ability to generate more accurate pseudo-labels not only for small objects such as pedestrians and cyclists but also for larger objects like cars and buses. This result clearly indicates that teacher networks specialized for each category can generate high-accuracy pseudo-labels, leading to significantly enhanced 3D object detection capabilities in the student network trained with these reliable pseudo-labels.

\section{Conclusion}\label{sec5}
In traditional object detection methods, there are limitations in improving detection performance when relying solely on labeled data. Consequently, recent approaches have focused on enhancing object detection performance by combining small amounts of labeled data with large-scale unlabeled data, leading to the development of 3D SSOD techniques. In this paper, we introduce MultipleTeachers, a 3D SSOD framework that creates optimized multiple teachers network for various categories. These multiple teachers collaborate to effectively guide a student network, resulting in robust detection capabilities. In addition, by utilizing high-quality pseudo-labels generated from professional teachers for vehicles, pedestrians, and cyclists, the learning efficiency of the student network is maximized, significantly enhancing object detection performance. Therefore, this approach effectively addresses the issue of low-quality pseudo-labels that often arises in traditional single teacher network paradigms. In addition, to mitigate issues caused by the sparsity of point clouds, we propose a novel data augmentation method PieAug. The method enhances the generalization ability and significantly improves the detection performance of the student network, since this augmentation provides richer learning information for objects.

Furthermore, recognizing the necessity of building the dataset for autonomous driving to accelerate research, we construct a new dataset LiO. This dataset comprehensively reflects various urban road environments and ensures data balance across diverse classes, including cars, buses, trucks, other vehicles, pedestrians, motorcycles, and bicycles. Moreover, following our rigorous guidelines, it is meticulously refined and annotated to build a high-quality labeled data.

To demonstrate the excellence and effectiveness of MultipleTeachers, extensive experiments are conducted on three large-scale autonomous driving datasets: WOD, KITTI, and LiO. Through these experiments, our proposed detector exhibited state-of-the-art performance, surpassing existing 3D SSOD models on all three datasets, thereby proving the superiority of our detection method.








\bibliography{sn-bibliography}

\end{document}